\documentclass{article}

\usepackage[numbers,square]{natbib}
\usepackage[nonatbib,final]{neurips_2022}

\usepackage[utf8]{inputenc} % allow utf-8 input
\usepackage[T1]{fontenc}    % use 8-bit T1 fonts
\usepackage{hyperref}       % hyperlinks
\usepackage{url}            % simple URL typesetting
\usepackage{booktabs}       % professional-quality tables
\usepackage{amsfonts}       % blackboard math symbols
\usepackage{nicefrac}       % compact symbols for 1/2, etc.
\usepackage{microtype}      % microtypography
\usepackage{xcolor}         % colors
\usepackage{todonotes}      % todo notes
\usepackage{booktabs}
\usepackage{multirow}
\usepackage{array}
\usepackage{wrapfig}
\usepackage{adjustbox}
\usepackage{graphicx}
\usepackage[ruled,vlined]{algorithm2e}
\usepackage[noend]{algorithmic}
\usepackage{amsmath,amssymb,amsthm}
\usepackage[title]{appendix}
\usepackage{float}
\newcommand\NoThen{\renewcommand\algorithmicthen{}}

\title{Learning-based \textcolor{black}{Motion} Planning in Dynamic Environments Using GNNs and Temporal Encoding}

\author{Ruipeng Zhang\\UCSD \And Chenning Yu\\UCSD \And Jingkai Chen\\MIT \And Chuchu Fan\\MIT \And Sicun Gao\\UCSD}

\begin{document}

\maketitle

\begin{abstract}
Learning-based methods have shown promising performance for accelerating motion planning, but mostly in the setting of static environments. For the more challenging problem of planning in dynamic environments, such as multi-arm assembly tasks and human-robot interaction, motion planners need to consider the trajectories of the dynamic obstacles and reason about temporal-spatial interactions in very large state spaces. We propose a GNN-based approach that uses temporal encoding and imitation learning with data aggregation for learning both the embeddings and the edge prioritization policies. Experiments show that the proposed methods can significantly accelerate online planning over state-of-the-art complete dynamic planning algorithms. The learned models can often reduce costly collision checking operations by more than 1000x, and thus accelerating planning by up to 95\%, while achieving high success rates on hard instances as well. 
\end{abstract}

\section{Introduction}

Motion planning for manipulation has been a longstanding challenge in robotics~\cite{manip1,manip2}. Learning-based approaches can exploit patterns in the configuration space to accelerate planning with promising performance~\cite{GVIN,Fastron,Clearance}. Existing learning-based approaches typically combine reinforcement learning (RL) and imitation learning (IL) to learn policies for sampling or ranking the options at each step of the planning process~\cite{MPNet,Implicit,NEXT}. Graph Neural Networks (GNNs) are a popular choice of representation for motion planning problems, because of their capability to capture geometric information and are invariant to the permutations of the sampled graph~\cite{GNNMP,decentral,attentionqingbiao,chenning21}. 

\begin{figure}[t!]
  \centering
    \includegraphics[width=0.8\textwidth]{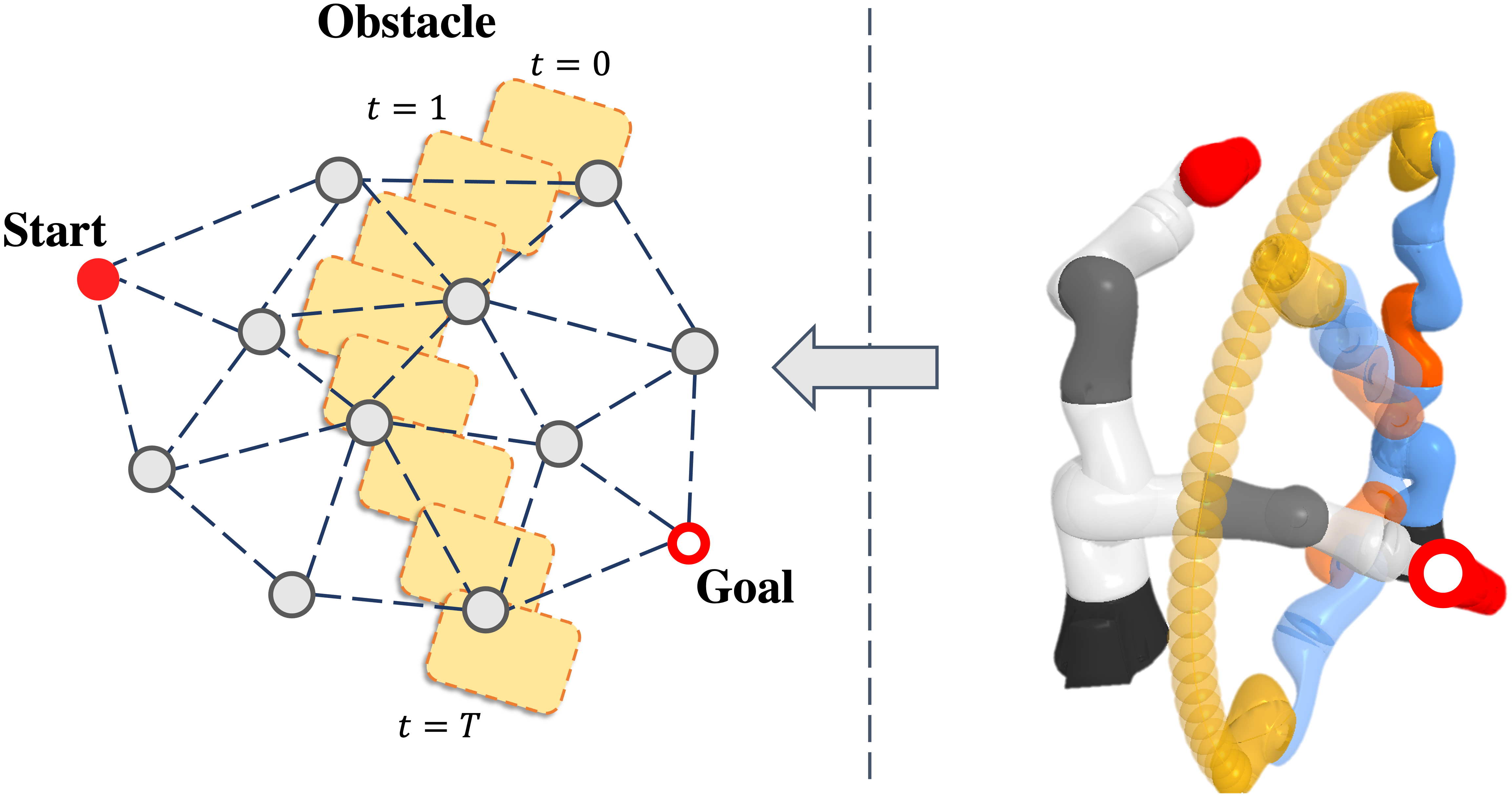}
    \vspace{-5pt}
    \caption{Left: A sampled graph from the configuration space. A dynamic obstacle, colored in yellow, moves over time from $t=0$ to $t=T$. The goal of our approach is to search for a path on the graph connecting the start to the goal, without collision with the obstacle at any timestep. Right: A successful plan where the ego-robot (grey arm) avoids collision with the dynamic obstacle (blue arm) and reaches the goal.}
    \label{fig:intro}
\end{figure}

\textcolor{black}{Motion} planning in dynamic environments, such as for multi-arm assembly and human-robot interaction, is significantly more challenging than in static environments. 
Dynamic obstacles produce trajectories in the temporal-spatial space, so the motion planner needs to consider global geometric constraints in the configuration space at each time step (Figure~\ref{fig:intro}). This dynamic nature of the environment generates the much larger space of sequences of graphs for sampling and learning, and it is also very sensitive to the changes in one single dimension: time. A small change in the timing of the ego-robot or the obstacles in two spatially similar patterns may result in completely different planning problems. For instance, the dynamic obstacle may create a small time window for the ego-robot to pass through, and if that window is missed, then the topology of configuration space can completely change. Consequently, we need to design special architectures that can not only encode the graph structures well, but also infer temporal information robustly. 
Indeed, complete search algorithms for dynamic motion planning, such as the leading method of Safe Interval Path Planning (SIPP) and its variations~\cite{phillips2011sipp,li2019safe,gonzalez2012using,narayanan2012anytime}, 
focus on reasoning about the temporal intervals that are safe for the ego-robot. These complete algorithms typically require significantly more computation and collision checking operations compared to the static setting. As it is proved in \cite{reif1994motion}, the computational complexity of planning with the moving obstacles is NP-hard even when the ego-robot has only a small and fixed number of degrees of freedom of movement. 

We propose a novel Graph Neural Network (GNN) architecture and the corresponding training algorithms \textcolor{black}{for motion} planning in dynamic environments. We follow the framework of sampling-based motion planning~\cite{DBLP:journals/trob/KavrakiSLO96,RRT*}, where path planning is performed on random graphs sampled from the configuration space. The GNN takes in the following inputs: the sampled graph in the configuration space, the obstacle's trajectory in the workspace, and the current state of the ego-robot. The output is a vector of priority values on the candidate edges at the current state of the ego-robot. The encoding is performed in two stages. In the first stage, we encode the graph structure using attention mechanisms~\cite{Attention}, and also design a temporal encoding approach for the obstacle trajectories. The temporal encoding uses the idea of positional encoding from the Transformer and NeRF~\cite{Attention,Nerf}, which encourages the neural network to capture temporal patterns from high-frequency input signals. In the second stage of encoding, we incorporate the ego-robot's current vertex in the configuration space, the local graph structure, and the current time-shifted trajectories of the obstacles. This two-stage structure extends the previous use of GNNs in static environments~\cite{decentral,attentionqingbiao,chenning21} and it is important for making high-quality predictions on the priority values. The entire GNN of both stages will be trained simultaneously in an end-to-end fashion. Due to the complexity of the GNN architecture, we observe that RL-based approaches can hardly train generalizable models based on the architecture, and using imitation learning with data aggregation (DAgger) is the key to good performance. We utilize SIPP as the expert, first perform behavior cloning as warm-up, and then allow the ego-robot to self-explore and learn from the expert following the DAgger approach~\cite{DAgger,decentralarm}.

We evaluate the proposed approach in various challenging dynamic motion planning environments ranging from 2-DoF to 7-DoF KUKA arms. Experiments show that our method can significantly reduce collision checking, often by more than 1000x compared to the complete algorithms, which leads to reducing the online computation time by over 95\%. The proposed methods also achieve high success rates in hard instances and consistently outperform other learning and heuristic baselines. 

\section{Related Work}

\noindent{\bf \textcolor{black}{Motion} Planning in Dynamic Environments.} 
Planning in dynamic environment is fundamentally difficult~\cite{reif1994motion}. Safe Interval Path Planning (SIPP)~\cite{phillips2011sipp} and its variations~\cite{gonzalez2012using,narayanan2012anytime,li2019safe} are the leading framework for \textcolor{black}{motion} planning in dynamic environments. The method significantly reduces the temporal-spatial search space by grouping safe configurations over a period of time \textit{safe intervals}, \cite{phillips2011sipp}. Based on this data structure, it can find the optimal paths on a configuration graph. Further improvements on SIPP include leveraging state dominance \cite{gonzalez2012using} and achieving anytime search~\cite{narayanan2012anytime}. 

Recently, \cite{li2019safe} shows SIPP is also powerful in planning the paths of high-dimensional manipulators on configuration roadmaps \cite{DBLP:journals/trob/KavrakiSLO96} with the presence of dynamic obstacles, which is the problem of interests in this paper. Other methods have been proposed for path planning in dynamic environments for mobile robots specifically~\cite{hauser2012responsiveness, park2012itomp,dong2016motion}. Such problems are typically more efficiently solvable because of the low-dimensional configuration space and the use of more conservative planning. 

\noindent{\bf Learning-based Motion Planning.} 
Learning-based approaches typically consider motion planning as a sequential decision-making problem that can be tackled with reinforcement learning or imitation learning. With model-based reinforcement learning, DDPG-MP~\cite{RSS} integrates the known dynamic of robots and trains a policy network. \cite{Obstacle} improves obstacle encoding with the position and normal vectors. OracleNet~\cite{OracleNet} learns via oracle imitation and encodes the trajectory history by an LSTM~\cite{LSTM}. Other than greedily predicting nodes that sequentially form a trajectory, various approaches have been designed to first learn to sample vertices, and then apply search algorithms on the sampled graphs~\cite{L2RRT}. \cite{CVAE,MPNet,Implicit} learn sampling distributions to generate critical vertices from configuration space. \cite{Adaptive,NEXT,GPPN} design neural networks that can better handle structured inputs. Learning-based approaches have also been proposed to improve collision detection~\cite{Fastron,Clearance,chenning21} and for exploration of edges on fixed graphs~\cite{VIN,GPPN}. For non-static environments, most works focus on multi-agent scenarios \cite{GNNCTRM,decentralarm} that do not involve non-cooperative dynamic obstacles. 

\noindent{\bf Graph Neural Networks for Motion Planning.}  Graph neural networks are permutationally invariant to node ordering, which becomes a natural choice for learning patterns on graphs. For motion planning, \cite{GNNMP} utilizes GNN to identify critical samples. \cite{decentral,attentionqingbiao,GamaLTPR22} predicts the action for each agent in the grid-world environment using graph neural networks.
\cite{vijayswarm,vijaygpg} learns the control policy for each robot in a large-scale swarm.~\cite{zhou2021graph} learns the submodular action selection for continuous space.~\cite{chenning21} learns a GNN-based heuristic function for search in static environment with completeness guarantee.

\section{Preliminaries}

\noindent{\bf Sampling-based Motion Planning with Dynamic Obstacles.} We focus on the sampling-based motion planning, in which a random graph is formed over samples from the {\em configuration space} $C\subseteq \mathbb{R}^n$ where $n$ is the number of degree-of-freedom for the ego-robot. The sampled vertex set $V$ always contains the start vertex $v_s$ and goal vertex $v_g$. The edges in the graph $G=\langle V,E\rangle$ are determined by r-disc or k-nearest-neighbor (k-NN) rules~\cite{rdisc,k-NN}. 
We assume global knowledge of the trajectories of the dynamic obstacles. We represent the trajectories of dynamic obstacles in the {\em workspace} as the vector of all the joint positions in the time window of length $T>0$.
The goal of the motion planning problem is to find a path from $v_s$ to $v_g$ in the sampled graph that is free of collision with the dynamic obstacles at all time steps in $[0,T]$.

\noindent{\bf Graph Neural Networks (GNNs).} GNNs learn representations over of vertices and edges on graphs by message passing. With MLP networks $f$ and $g$, GNN encodes the representation $h_i^{\textcolor{black}{(k+1)}}$ of vertex $v_i$ after $k$ aggregation steps defined as
\begin{equation}\label{eq2}
\left.\begin{aligned}
h_i^{(k+1)} = g(h_i^{(k)},  \textcolor{black}{\oplus}(\left\{f(h_i^{(k)}, h_j^{(k)})\mid (v_i,v_j)\in E\right\})) 
\end{aligned}\right.
\end{equation}
where $h_i^{(1)}=x_i$ can be some arbitrary vector of initial data for the vertex $v_i$.
$\oplus$ is typically some permutation-invariant aggregation function on sets, such as mean, max, or sum. 
We use the attention mechanism to encode the obstacle features. 
In the general form of the attention mechanism, there are $n$ keys, each with dimension $d_k$: $K\in \mathbb{R}^{n\times d_k}$, each key \textcolor{black}{has} a value $V \in \mathbb{R}^{n\times d_v}$. Given $m$ query vectors $Q \in \mathbb{R}^{m\times d_k}$, we use a typical attention function $\mathbf{Att}(K, Q, V)$ for each query as
$\mathbf{Att}(K, Q, V)=\text{softmax}(QK^T/\sqrt{d_k})V$ \cite{Attention}. 

\noindent{\bf Imitation Learning.} Imitation learning aims to provide guidance to train policy without explicitly designing reward functions. 
Given a distribution of the oracle actions $\pi_{oracle}$, it tries to learn a new policy distribution $\pi$ that minimize the deviation from the oracle, i.e.
$
    \pi^* = \operatorname*{argmin}_{\pi}{D(\pi,\pi_{oracle})}
$
, where $D$ is the difference function between the two distributions, which can be represented as $p$-norm or $f$-divergence. We use imitation learning to train our GNN models from the demonstration of the oracle planner. Specifically, in the task of sampling-based motion planning, the oracle predicts the priority values of subsequent edges and prioritize the one given by the oracle. However, imitation learning based on behavior cloning often suffers from distribution drift, which can be mitigated by imitation with data aggregation (DAgger)~\cite{DAgger}. With DAgger, the learner can actively query the oracle on states that are not typically provided in the expert trajectories to robustify the learned policy. 

\section{GNN-TE: Dynamic Motion Planning with GNNs and Temporal Encoding}

\subsection{Overall Architecture}

\begin{figure}[t!]
    \centering
    \includegraphics[width=0.9\textwidth]{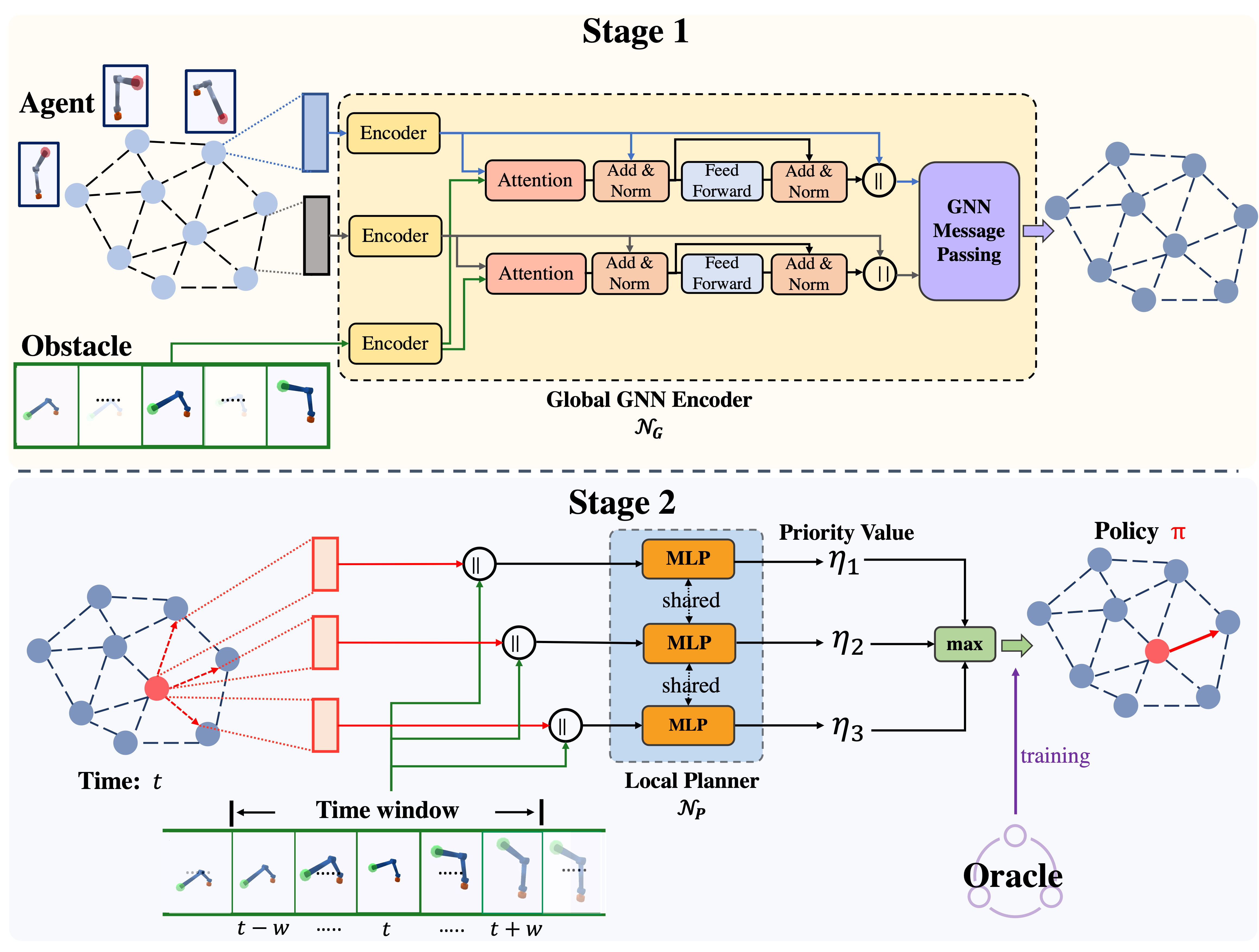}
    \caption{The overall two-stage architecture of the proposed \textbf{GNN-TE}. In Stage 1, we encode global information of the ego-arm and the obstacles, using attention mechanisms, and output the encoding of each edge. In Stage 2, the local planner will take in the output from Stage 1 along with the obstacle encoding within the relevant time window, to predict the priority value of each outgoing edge. The planner will propose the edge with the highest priority value to take as the output policy.}
    \label{fig:gnn-arch}
    \vspace{-5pt}
\end{figure}

We design the dynamic motion planning network \textbf{GNN-TE} to capture the spatial-temporal nature of dynamic motion planning. The forward pass in the network consists of two stages. The first stage is the global GNN encoder $\mathcal{N}_G$ that encodes the global information of the ego-robot and obstacles in the environment. The second stage is the local planner $\mathcal{N}_p$ that assigns priority on edges, utilizing the encoding output of the first stage. Fig \ref{fig:gnn-arch} shows the overall two-stage architecture. 

\noindent{\bf First-stage global GNN encoder $\mathcal{N}_G$}.
The GNN encoder $\mathcal{N}_G$ takes in a sampled random geometric graph $G=\langle V,E\rangle$, $V = \{v_s, v_g, v\}$. For an $n$-dimensional configuration space, each vertex $v_i\in \mathbb{R}^{n+1}$ contains an $n$-dimensional configuration component and a $1$-dimensional one-hot label indicating if it is the special goal vertex.

The vertices and the edges are first encoded into a latent space with $x\in \mathbb{R}^{|V|\times d_h}, y\in \mathbb{R}^{|E|\times d_h}$, where $d_h$ is the size of the encoding. \textcolor{black}{Specifically, %to encode the graph while informing the network of the task specific goal, 
to get the feature $x_i$ for the $i$-th node $v_i\in V$, we use $
    x_i=g_x(v_i, v_g, v_i-v_g, ||v_i-v_g||^2_2)$. To get the feature $y_l$ for the $l$-th edge $e_l:\langle v_i,v_j\rangle \in E$ , we use
    $y_l=g_y(v_i, v_j, v_j-v_i)$.
The $g_x$ and $g_y$ are two different two-layer MLPs.} The L2 distance to the goal $||v-v_g||^2_2$ serves as the heuristic information for $\mathcal{N}_G$.

The dynamic obstacles $O$ form barriers on the top of the graph $G$, and we incorporate their trajectories to $\mathcal{N}_G$ to leverage the global environment and assist the planning of $\mathcal{N}_P$. Additionally, to inform the networks about the relative time over the horizon, we incorporate {\em temporal encoding} with the obstacles. Given the obstacle position $O_t$ at time step $t$ and a two-layer MLP $g_o$, the obstacle is encoded as $\mathcal{O}_t = g_o(O_t)+TE(t)$, which adds $g_o(O_t)$ and $TE(t)$ element-wisely. $TE(t)$ is the temporal encoding at time step $t$, which we will discuss at Section \ref{TESection}. 
With the sequential property of the trajectory, we use the attention mechanism to model the temporal-spatial interactions between the ego-arm and the obstacles. Concretely, the obstacles are encoded into the vertex and edge of $G$ as:
\begin{align}
    x &= x + \mathbf{Att}(f_{K_x^{(i)}}(\textcolor{black}{\mathcal{O}}), f_{Q_x^{(i)}}(x), f_{V_x^{(i)}}(\textcolor{black}{\mathcal{O}})) \\ 
    y &= y + \mathbf{Att}(f_{K_y^{(i)}}(\textcolor{black}{\mathcal{O}}), f_{Q_y^{(i)}}(y), f_{V_y^{(i)}}(\textcolor{black}{\mathcal{O}}))
    \label{eq:attention}
\end{align}

Taking the vertex and edge encoding $x, y$, the GNN $\mathcal{N}_G$ aggregates the local information for each vertex and edge from the neighbors with the following operation with 2 two-layer MLPs $f_x$ and $f_y$:
\begin{equation}
\begin{split}
x_i&=\max \left(x_i, \max\{f_x(x_j-x_i,x_j, x_i,y_l)\mid e_l:\langle v_i,v_j\rangle \in E\}\right), \forall v_i \in V\\
y_l&= \max(y_l, f_y(x_j-x_i,x_j, x_i)), \forall e_l:\langle v_i,v_j\rangle \in E
\label{coreupdate}
\end{split}
\end{equation}
Note that here we use $\max$ as the aggregation operator to gather the local geometric information, due to its empirical robustness to achieve the order invariance~\cite{PointNet}. The edge information is also incorporated into the vertex by adding $y_l$ as the input to $f_x$. Also, because Equation~\ref{coreupdate} is a homogeneous function that updates on the $x$ and $y$ in a self-iterating way, we can update without introducing redundant layers over multiple loops. After several iterations, the first-stage $\mathcal{N}_G$ outputs the encoding of each vertex $x_i$ and edge $y_l$.

\noindent{\bf Second-stage local planner $\mathcal{N}_P$}.
After $\mathcal{N}_G$ encodes the information of the configuration graph and obstacles, the second-stage local planner $\mathcal{N}_P$ utilizes the encoding and performs motion planning. Specifically, when arriving at a vertex $v_i$ at time $t_i$, $\mathcal{N}_P$ predicts the priority value $\eta_{e_i}$ of all the connected edges $e_i\in \textcolor{black}{E_i}$ with the expression $\eta_{e_i} = f_p(y_{e_i}, \textcolor{black}{\mathcal{O}}_{t_i-w}, \textcolor{black}{\mathcal{O}}_{t_i-w+1},...,\textcolor{black}{\mathcal{O}}_{t_i+w-1}, \textcolor{black}{\mathcal{O}}_{t_i+w})$, where $f_p$ is an MLP. Note that in addition to the encoding of the connected edges, we also input the local obstacle encoding within a time window $w$ of the current arrival time $t_i$. This provides local information for $\mathcal{N}_P$ to plan towards the goal vertex, while considering the barriers of dynamic obstacles to avoid collisions. At inference time, we use $\mathcal{N}_P$ to choose the edge with the highest priority value while keeping track of the current time.

\subsection{Temporal Encoding}\label{TESection}

\begin{figure}[h]
    % \makebox[\textwidth][c]
    {\includegraphics[width=0.9\textwidth]{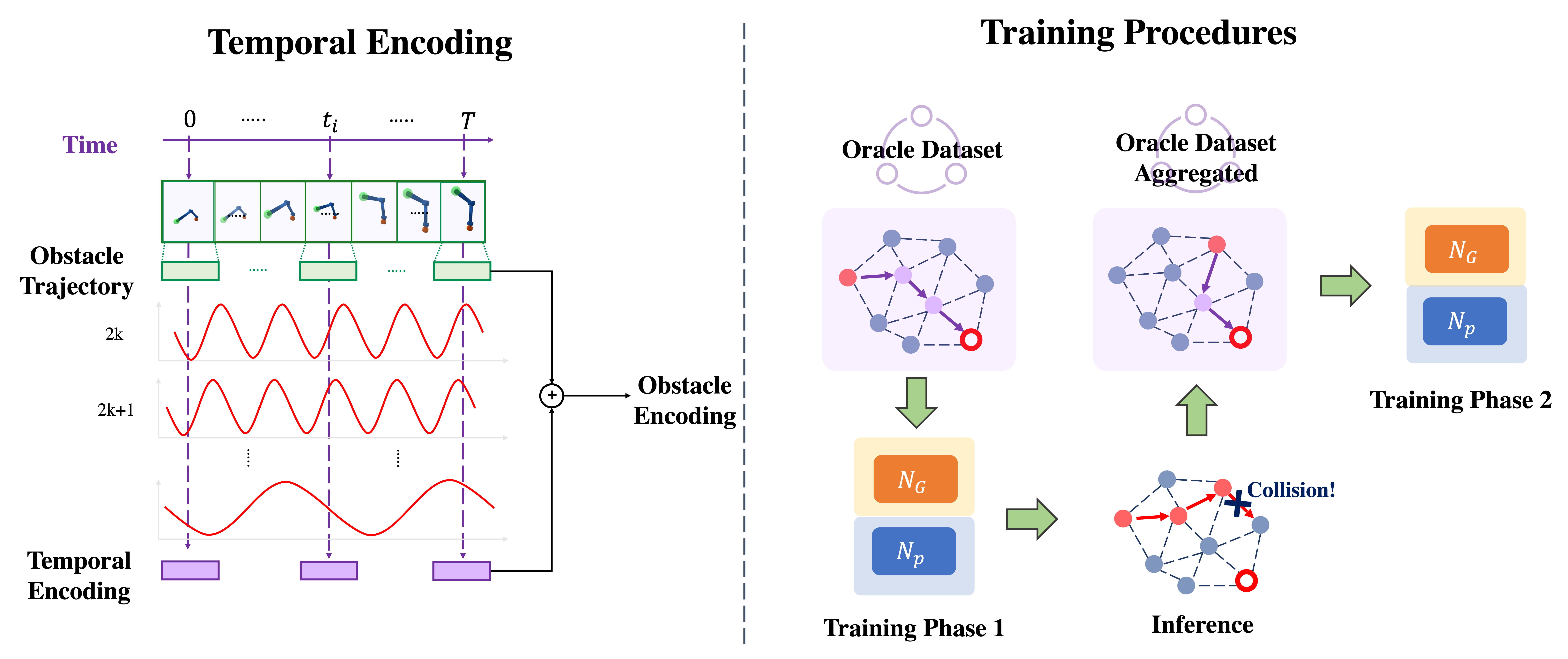}}
    \caption{Left: Temporal encoding is incorporated when representing the dynamic obstacle sequence. Right: Training procedures with DAgger. The proposed GNN is first trained to imitate an optimal oracle, then improves itself by self-exploring the data with feedback from the oracle.}
    \label{fig:temporal-encoding & training}
    \vspace{-10pt}
\end{figure}

Positional encoding is a crucial design in the Transformer architecture \cite{vaswani2017attention} for making use of the order of the sequence. Dynamic motion planning requires the models to infer the relative position of obstacles and how they interact with the ego-arm at each time step. So along with the positions of the obstacles in the workspace, we add temporal encoding \textcolor{black}{$TE(t) \in \mathbb{R}^{d_{TE}}$ at each time step $t\in[0,\cdots,T]$, where its $2k$-th and $2k+1$-th dimensions} are computed as
\begin{align}
    TE(t, 2k) = \sin({\omega^{-2k/{d_{TE}}}}t)\ \mbox{ and }\ 
    TE(t, 2k+1) = \cos({\omega^{-2k/{d_{TE}}}}t)
    \label{eq:TE}
\end{align}
where $\omega\in \mathbb{Z}^+$ is a fixed frequency. We select the temporal encoding to have the same dimension as the obstacle input, and add them as the obstacle encoding before inputting into the networks $\mathcal{N}_G$ and $\mathcal{N}_P$. We illustrate the overall encoding procedure on the left of Figure~\ref{fig:temporal-encoding & training}.
\subsection{Training and Inference Procedures}

In each training problem, along with the dynamic obstacles $\mathcal{O}$, start vertex $v_s$ and goal vertex $v_g$, we sample a k-NN graph $G=\langle V,E\rangle, V=\{v_s, v_g, v\}$, where $v$ is the vertices sampled from the configuration space of the ego-arm. In the first stage, we use the global GNN encoder $\mathcal{N}_G$ to encode the graph and dynamic obstacles, and the local planner $\mathcal{N}_p$ in the second stage uses the encoding as the input to predict the priority value $\eta$ of the subsequent edges.

\noindent{\bf Imitation from SIPP with Data Aggregation.}
We train our two-stage network $\mathcal{N}_G$ and $\mathcal{N}_p$ in an end-to-end manner by imitating an oracle. Specifically, we use Safe Interval Path Planning (SIPP) \cite{phillips2011sipp} to compute the shortest non-collision motion path and use it as the oracle.

In the first stage, $\mathcal{N}_G$ will process the graph and the obstacle trajectories, then output the encoded features of each vertex and edge. Then in the second stage, we train the networks to imitate the oracle SIPP along the optimal path. Concretely, starting from the $v_s$, SIPP provides the optimal path $\pi^*=\{(v^*_i, t^*_i)\}_{i\in[0,n]}$ with the vertex $v^*_i$ and the corresponding arrival time $t^*_i$. When arriving at the vertex $v^*_i$ at time $t^*_i$, the local planner $\mathcal{N}_p$ will take in the edge feature $y_{e_i}$ along with the obstacle encoding in the time window $[t^*_{i-w}, t^*_{i+w}]$ to predict the priority value of all the subsequent edges $E_i$. Then it prioritizes the next edge on the optimal path $e^*_i:\langle(v^*_i, t^*_i), (v^*_{i+1}, t^*_{i+1})\rangle$ among $E_i$. We maximize the priority value $\eta_{e^*_i}$ of $e^*_i$ over all other $e_i \in E_i\setminus \{e^*_i\}$ with the standard cross entropy loss $L_{i} = -\log (\exp({\eta_{e^*_i}})/(\Sigma_{e_i \in E_i}\exp({\eta_{e_i}})))$. 

Since SIPP only provides the guidance on the optimal path, when the planned path given by $\mathcal{N}_p$ deviates from the optimal path, our network cannot sufficiently imitate the oracle. To this end, we use DAgger~\cite{DAgger} to encourage the network to learn from these sub-optimal paths. We first train our network for $k$ iterations with pure demonstrations from SIPP. Then we explore a path $\pi^k$ on the graph using the priority value predicted by the current network, which may not reach the goal vertex $v_g$ nor be optimal. We randomly stop at the vertex $v^k_i$ at time $t^k_i$ and query the oracle. SIPP treats $v^k_i$ and $t^k_i$ as the start vertex and the start time respectively, along with the obstacles trajectory starting at $t^k_i$, calculates the optimal path. The new demonstrations are aggregated to the previous dataset to keep training the network. The training procedures are showed on the right of Figure~\ref{fig:temporal-encoding & training}.

\noindent{\bf Inference with Time Tracking.} Given a graph $G=\langle V, E\rangle, V=\{v_s, v_g, v\}$, with the trajectories of obstacles $\mathcal{O}$, $\mathcal{N}_G$ will first encode the graph and obstacles. Next, $\mathcal{N}_P$ executes motion planning by predicting the priority values $\eta=\mathcal{N}_P(V,E_i,\mathcal{O}, \mathcal{N}_G, t_i)$ when arriving at vertex $v_i$ at time $t_i$, and follow the edge with the maximum one, i.e. $e_{\pi_i} = \operatorname*{argmax}_{e_i \in E_i}\eta_{e_i}$. After the edge $e_{\pi_i}$ is proposed by the network, we check the collision on $e_{\pi_i}$ on which the ego-arm starts moving at $t_i$. If there is no collision, we add $(e_{\pi_i}, t_i)$ into the current path $\pi$. Otherwise, we query $\mathcal{N}_P$ for another edge with the next greater priority value. The planning will end if we succeed in finding a path $\pi$ from $v_s$ to $v_g$ or fail when stopping at a vertex with all the connected edges with collisions. Optionally, when the latter one happens, we can backtrack to the last step and query for the top-k greatest priorities in turn. Further discussions are covered in the experiments section.

\section{Experiments}

\begin{figure}[!th]
  \centering
  \begin{adjustbox}{width=1\textwidth,center=\textwidth}
    \includegraphics[]{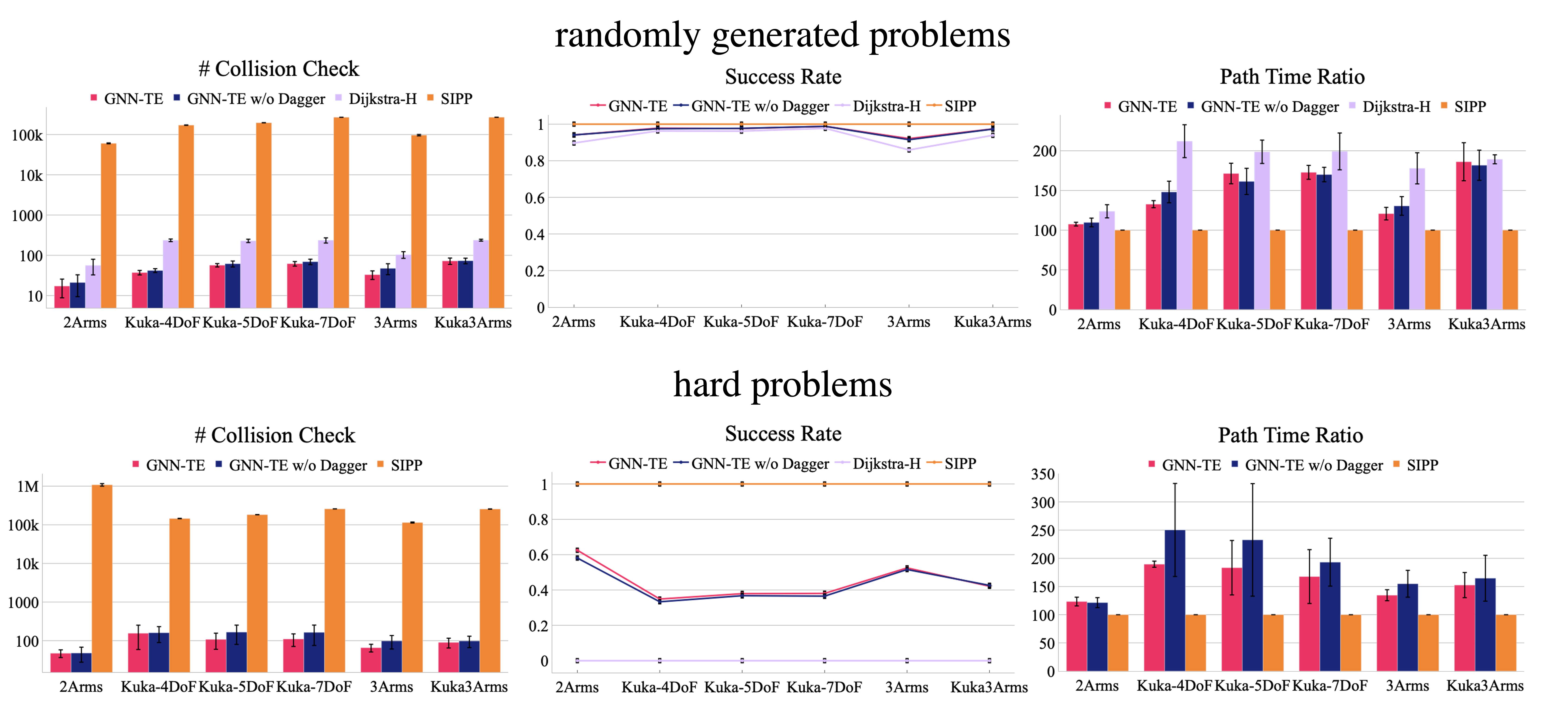}
    \end{adjustbox}
    \caption{Overall Performance of baselines on collision checking, success rate, and path time ratio. Our approach significantly reduces collision
checking more than 1000x compared to the complete algorithm SIPP, improves the overall planning efficiency, and achieves high success rates.}
    \label{fig:overall-performance}
\end{figure}

\noindent{\bf Experiment Setup.} We evaluate the proposed methods on various multi-arm assembly tasks where we manipulate the ego-arm and avoid collisions with other obstacles, including moving arms and static obstacles. Specifically, the environments cover arms of 2 to 7 Degree of Freedom and 1 to 3 moving obstacle arms, including (i) 2Arms: 1 obstacle arm with 2 DoF. (ii) Kuka-4DoF: 1 obstacle arm with 4 DoF. (iii) Kuka-5DoF: 1 obstacle arm with 5 DoF. (iv) Kuka-7DoF: 1 obstacle arm with 7 DoF. (v) 3Arms: 2 obstacle arms with 2 DoF. (vi) Kuka3Arms: 2 obstacle arms with 7 DoF. Without loss of generality, we assume all the robot arms in an environment are of the same DoF and move at the same and constant speed. The static obstacles are represented as cuboids in the experiments.

\noindent{\bf Baselines.} We compare our method \textbf{GNN-TE} with the lazy sampling-based dynamic motion planning method \textbf{Dijkstra-Heuristic} (\textbf{Dijkstra-H}). It prioritizes the edges based on the shortest distance to the goal on the configuration graph at each time step. We also compare our method with the oracle \textbf{SIPP} \cite{phillips2011sipp}, which is a search algorithm that generates optimal paths using safe intervals.

\noindent{\bf Datasets.} We randomly generate 2000 problems for training and 1000 problems for testing. In each problem, we randomly generate fixed-length trajectories of the moving obstacle arms and sample 1000 vertices in the configuration space of the ego-arm that are collision-free with the static obstacles, then provide them to the oracle to generate demonstration trajectories. In order to further investigate the performance of algorithms in challenging environments, we also generate 1000 hard problems, where \textbf{Dijkstra-H} fails to find feasible paths. 

\noindent{\bf Training and Testing Details.}
We first train the \textbf{GNN-TE} on all the training problems for 200 epochs. Afterward, we generate 1000 new training data with DAgger, and trained for another 100 epochs. We test all the algorithms on the provided graphs. We randomly split the test problems into 5 groups, and calculate the performance variance.

\noindent{\bf Evaluation Metrics.} We measure the average number of \textbf{collision checking} of the common success cases among all the algorithms to evaluate the effectiveness in avoiding obstacles. Collision checking is the more expensive computation in motion planning, and reducing it can lead to a significant acceleration of online computation. We also report the average \textbf{success rate} over all the testing problems for all the algorithms. A trajectory is successful, only if the ego-arm does not collide with any obstacles, and eventually reaches the goal given a limited time horizon. We also measure the optimality of the methods by comparing the \textbf{path time ratio} to the optimal paths found by SIPP on each success case. Note that the path time metric only measures the cost of the planned path, and does not include the online computation time needed. 

\subsection{Overall Performance}

Figure \ref{fig:overall-performance} shows the overall performance of the algorithms in various environments, including all the baselines and \textbf{GNN-TE} without DAgger, i.e., pure imitation learning from the oracle. 

In all the environments, \textbf{SIPP} gives the optimal complete non-collision path, but it suffers from the excessive amount of collision checking. \textbf{GNN-TE} significantly reduces the collision checking by more than 1000x, which corresponds to reducing online computation time by over 95\%. At the same time, the methods have high success rates and a better path time ratio compared to simpler heuristics.

\begin{wrapfigure}{r}{0.5\textwidth}
    \includegraphics[width=0.5\textwidth]{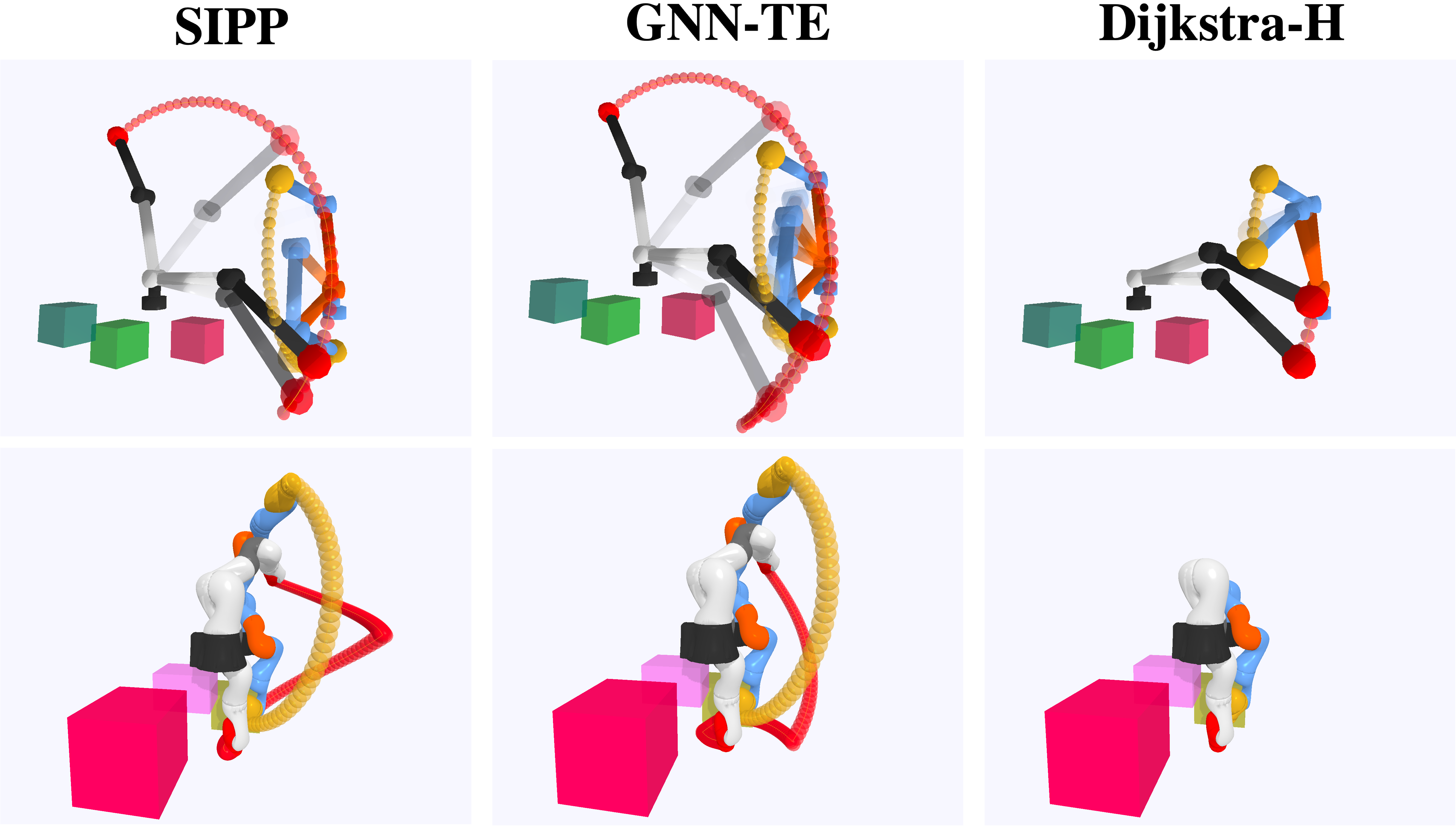}
    \caption{Snapshots of the trajectories of the output path on 2 test cases. The environments are 2Arms (first row) and Kuka-7DoF (second row).
The ego-arm is black and white with a red end-effector. The obstacle arm is blue and orange with a yellow end-effector. In both environments, \textbf{Dijkstra-H} fails to find a path, while our method can yield a near-optimal path compared to the oracle \textbf{SIPP}.}
\vspace{-0.3cm}
    \label{fig:overall-snapshot}
\end{wrapfigure}

Fig. \ref{fig:overall-snapshot} shows the performance snapshots of the algorithms on a 2Arms and a Kuka-7DoF test case. In both cases, our method successes in planning a near-optimal path compared to the oracle \textbf{SIPP} whereas \textbf{Dijkstra-H} fails.

\noindent{\bf Performance on randomly generated test problems.}
As shown in Fig.\ref{fig:overall-performance}, our methods significantly reduces the collision checking
by over 1000x compared to \textbf{SIPP}, i.e., 60k, 97k, 171k, 197k, 269k, 269k to 17.24, 33.08, 37.36, 56.89,  61.99, 72.63 on 2Arms, 3Arms, Kuka-4DoF, Kuka-5DoF, Kuka-7DoF, Kuka3Arms respectively. 
Because \textbf{SIPP} needs to search in both space and time dimensions, it suffers from a large number of collision checking.
Our approach benefits from learning and achieves a much less amount of collision checking even in the dynamic setting while not compromising much of the success rate, in which case our method outperforms \textbf{Dijkstra-H} in all the environments with higher success rates.
Note that as a crucial part of the training procedures, DAgger also improves performance in all environments. 

\noindent{\bf Performance on hard test problems.}
On the hard test problems, \textbf{Dijkstra-H} fails to find feasible paths in the dynamic environment. Comparatively, our \textbf{GNN-TE} can successfully find solutions to these problems with considerable success rates and acceptable path time ratios. It is also worth noting that DAgger can better assist in improving the performance in the more challenging scenarios compared to the randomly generated problems.

\noindent{\bf Optional backtracking search.} 
In the Appendix \ref{AppendixBT}, we also report the result of \textbf{GNN-TE} and \textbf{Dijkstra-H} and them with backtracking \textbf{GNN-TE w. BT} and \textbf{Dijkstra-H w. BT} on 2Arms. On the same algorithm, the optional backtracking search will only result in a higher success rate while not affecting the path time ratio and collision checking on the common success cases. The result of \textbf{Dijkstra-H w. BT} shows that although it improves the success rate significantly but sacrificing a tremendous number of collision checking. Nevertheless, our method outperforms the heuristic method both with or without the backtracking search.

\noindent{\bf Comparison with end-to-end RL.} We also compare our approach with RL approaches, including DQN~\cite{DQN} and PPO~\cite{PPO}. The neural network architectures for these two baselines are implemented with the same GNN architectures as ours. We observe that even in 2Arms, the average success rate of DQN on the training set is only around 54\%, while PPO only has a success rate of around 20\%. They fail to find plans in test cases. We provide more details in the Appendix \ref{AppendixRL}.

\noindent{\bf Comparison with OracleNet-D.} We compare \textbf{GNN-TE} with a learning-based approach \textbf{OracleNet-D}, by modifing \textbf{OracleNet}\cite{OracleNet} to the dynamic version. We observe that the performance of \textbf{OracleNet-D} falls behind \textbf{GNN-TE} largely on all the metrics both in random and hard problems. Details are provided in the Appendix \ref{AppendixOracle}.

\subsection{Ablation Studies}\label{ablation}
\begin{figure}[t!]
  \centering
  \begin{adjustbox}{width=1\textwidth,center=\textwidth}
    \includegraphics[]{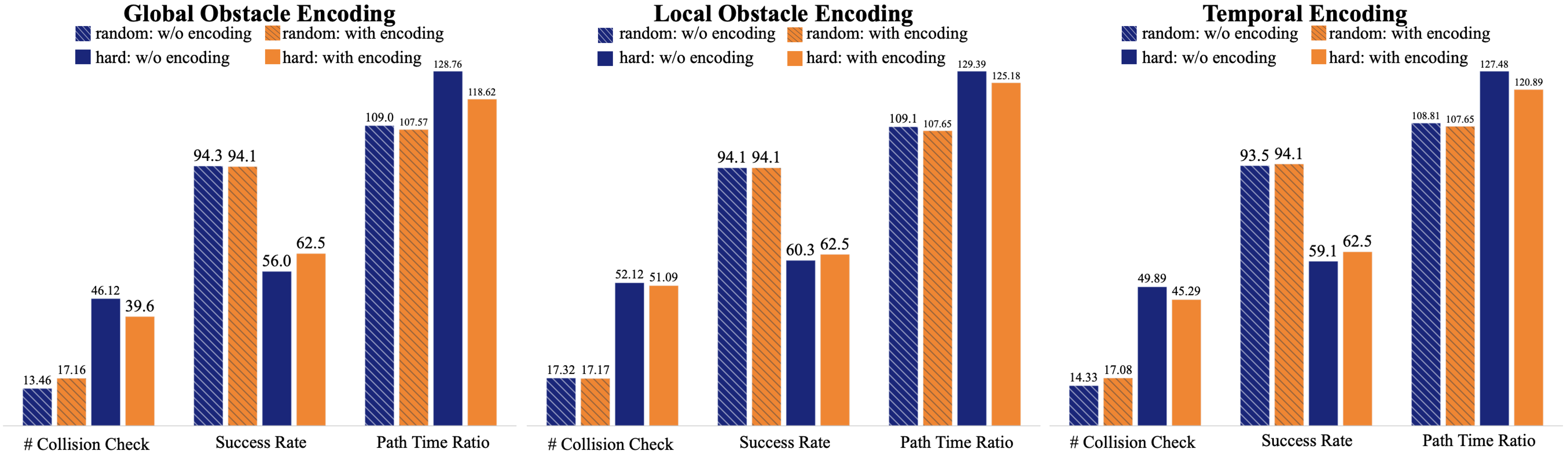}
    \end{adjustbox}
    \caption{Ablation studies on 2Arms of (1) global obstacle encoding, (2) local obstacle encoding, and (3) temporal encoding. We have demonstrated that all these three components improve the effectiveness of the proposed approach. See Section \ref{ablation} for more details.}
    \label{fig:ablation}
\end{figure}

We perform ablation studies on 2Arms of the different encoding in our model, including global and local obstacle encoding and temporal encoding. The results are shown in Fig. \ref{fig:ablation}.

\textbf{Global Obstacle Encoding.}
In stage 1, we encode both the configuration graph and the global obstacle trajectories with the GNN-based global encoder. To investigate the effectiveness of leveraging global obstacle information, we conduct an experiment in which we input the sampled configurations in stage 1 and only introduce the obstacle information in stage 2. As we can observe in the figure that although 
on random problems there are slight degrades of collision checking and success rate, the performance of path time ratio, success rate, and collision checking improves greatly in the hard environment. This results from the model receiving the overall trajectories of the obstacles, and the encoding helps to reduce collision checking and improves success rate, especially on complicated problems.

\textbf{Local Obstacle Encoding.}
We compare our model with the one omitting the local obstacle encoding in stage 2. We only input the temporal encoding corresponding to the arrival time to the local planner as the time indicator for planning. The result has shown that the local obstacle encoding within a time window directly helps the local planner perform better.

\textbf{Temporal Encoding.}
We analyze the importance of temporal encoding where we remove it from both two stages and only input the trajectory of obstacles in the models. The results also show that temporal information helps \textbf{GNN-TE} make use of the order of the trajectory sequence on both random and hard problems.

\section{Discussion and Conclusion}

We proposed a GNN-based neural architecture \textbf{GNN-TE} for motion planning in dynamic environments, 
where we formulate the spatial-temporal structure of dynamic planning problem 
with GNN and temporal encoding. 
We also use imitation learning with DAgger for learning both the embedding and edge prioritization policies. We evaluate the proposed approach in various environments, ranging from 2-DoF arms to 7-DoF KUKA arms. Experiments show that the proposed approach can reduce costly collision checking operations by more than 1000x and reduces online computation time by over 95\%. Future steps in this direction can involve using ideas from the complete planning algorithms, such as incorporating safe intervals, to improve the success rate on hard instances, as well as more compact architectures for further reducing online computation. 

\section{Acknowledgement}
This material is based on work supported by DARPA Contract No. FA8750-18-C-0092, AFOSR YIP FA9550-19-1-0041, NSF Career CCF 2047034, NSF CCF DASS 2217723, and Amazon Research Award. We appreciate the valuable feedback from Ya-Chien Chang, Milan Ganai, Chiaki Hirayama, Zhizhen Qin, Eric Yu, Hongzhan Yu, Yaoguang Zhai, and the anonymous reviewers.

\medskip

\bibliography{refs}

\newpage

\newpage
\begin{appendices}
\vspace{-50pt}
\section{Algorithms}

\vspace{-20pt}
\begin{algorithm}[!h]
   \caption{Stage1: Global GNN Encoder $\mathcal{N}_G$}
   \label{alg:gnn}
\begin{algorithmic}
   \STATE {\bfseries Input:} obstacles $O$, start $v_s$, goal $v_g$, network $g_x, g_y$, $K_x^{(i)}, Q_x^{(i)}, V_x^{(i)}, K_y^{(i)}, Q_y^{(i)}, V_y^{(i)}$ for $i$-th embedding dimension.
   \NoThen
   \STATE Sample $n$ nodes $v_1, \cdots, v_n$ from configuration space of ego-arm robot
   \STATE Initialize $G = \{V:\{v_s, v_g, v_1, \cdots, v_n\}, E:\text{k-NN}(V)\}$
   \STATE Initialize encoding of vertices and edges
   \vspace{-5pt}
   \begin{align*}
       x_i&=g_x(v_i, v_g, v_i-v_g, ||v_i-v_g||^2_2), \forall v_i \in V\\ 
        y_l&=g_y(v_i, v_j, v_j-v_i), \forall e_l:\langle v_i, v_j\rangle \in E
   \end{align*}
   \vspace{-15pt}
   \STATE Initialize obstacle encoding $\mathcal{O}_t=g_o(O_t)+TE(t), \forall t \in [0,\cdots, T]$ using Eq. \ref{eq:TE}
   \STATE Encode obstacles into vertices and edges using Eq. \ref{eq:attention}
   \STATE Message Passing using Eq. \ref{coreupdate}
   \RETURN Encoding of edges $\{y_l\}$
\end{algorithmic}
\end{algorithm}

\vspace{-20pt}
\begin{algorithm}[!h]
   \caption{Stage2: Lobal Planner $\mathcal{N}_P$}
   \label{alg:planner}
\begin{algorithmic}
   \STATE {\bfseries Input:} graph $G=\langle V, E\rangle$, encoding of edges $y_l$, obstacle encoding $\mathcal{O}$, time window $w$, global GNN encoder $\mathcal{N}_G$, local planner $\mathcal{N}_P$, goal-reaching constant $\delta$.
   \NoThen
   \STATE Initialize $i=0, v_0=v_s, t_0=0, \pi={(v_0, t_0)}, E_0=\{e:\langle v_s, v_k\rangle \in E, \forall v_k\in V\}$
   \REPEAT
   \STATE $\eta = \mathcal{N}_P(V,E_i,\mathcal{O}, \mathcal{N}_G, t_i)$
   \STATE select $e_j=\arg\max_{e_l\in E_i} \eta_l$, and $e_j$ connects $\langle v_i,v_j\rangle$
   \IF { $e_j$ is collision-free when start moving from $t_i$} 
        \STATE $t_{i+1} = t_{i} + \Delta(v_i, v_j)$ \tcp*{$\Delta(v_i, v_j)$ is the travel time from $v_i$ to $v_j$}
        \STATE $\pi_{i+1} \leftarrow \pi_{i} \cup\{(v_j,t_{i+1})\}$
        \STATE $v_{i+1} \leftarrow v_{j}$
        \STATE $E_{i+1}=\{e: \langle v_{i+1},v_k \rangle \in E, \forall v_k \in V\}$
        \IF{$||v_{i+1}-v_g||_2^2 \leq \delta$}
            \RETURN $\pi$
        \ENDIF
        \STATE $i\leftarrow i+1$
   \ELSE
        \STATE $E_{i} = E_{i} \setminus {e_j}$
   \ENDIF

   \UNTIL{$E_i=\emptyset$}
   \RETURN $\emptyset$
\end{algorithmic}
\end{algorithm}

\vspace{-15pt}
\begin{algorithm}[H]
   \caption{\textcolor{black}{Dijkstra-H}}
   \label{alg:dij-H}
\begin{algorithmic}
   \STATE {\bfseries Input:} graph $G=\langle V, E\rangle$, start $v_s$, goal $v_g$, goal-reaching constant $\delta$.
   \NoThen
   \STATE Sample $n$ nodes $v_1, \cdots, v_n$ from configuration space of ego-arm robot.
   \STATE Initialize $G = \{V:\{v_s, v_g, v_1, \cdots, v_n\}, E:\text{k-NN}(V)\}$
   \STATE Calculate the shortest distance $d_{v_k}$ on the graph from $v_g$ to each node $v_k\in V$ using Dijkstra's algorithm.
   \STATE Initialize $i=0, v_0=v_s, t_0=0,\pi={(v_0, t_0)}, E_0=\{e:\langle v_s, v_k\rangle \in E, \forall v_k \in V\}$
   \REPEAT
   \STATE select $v_j=\arg\min_{\langle v_i, v_j\rangle\in E_i} d_{v_j}$
   \IF {$\langle v_i, v_j\rangle$ is collision-free when start moving from $t_i$} 
        \STATE $t_{i+1} = t_{i} + \Delta(v_i, v_j)$ \tcp*{$\Delta(v_i, v_j)$ is the travel time from $v_i$ to $v_j$}
        \STATE $\pi_{i+1} \leftarrow \pi_{i} \cup\{(v_j,t_{i+1})\}$
        \STATE $v_{i+1} \leftarrow v_{j}$
        \STATE $E_{i+1}=\{e: \langle v_{i+1},v_k\rangle\in E,  \forall v_k \in V\}$
        \IF{$||v_{i+1}-v_g||_2^2 \leq \delta$}
            \RETURN $\pi$
        \ENDIF
        \STATE $i\leftarrow i+1$
   \ELSE
        \STATE $E_{i} = E_{i} \setminus {e_j}$
   \ENDIF

   \UNTIL{$E_i=\emptyset$}
   \RETURN $\emptyset$
\end{algorithmic}
\end{algorithm}

\section{\textcolor{black}{Network Architecture Details}}
We provide the numbers of network parameters in Table \ref{tab:netdetail}. Please refer to \ref{fig:gnn-arch} for the overall two-stage architecture of the proposed \textbf{GNN-TE}.

\begin{table}[H]
  \centering
  \caption{Network Architecture Details}
    \begin{tabular}{c|c}
    \toprule[2pt]
    \textbf{Name} & \textbf{Model} \\
    \hline
    \multicolumn{2}{c}{\textbf{Stage1 Global GNN Encoder}} \\
    \hline
    Node Encoder Net $g_x$ & MLP((config\_size+1)*4,32),MLP(32,32) \\
    \hline
    Edge Encoder Net $g_y$ & MLP((config\_size+1)*3,32),MLP(32,32) \\
    \hline
    Obstacle Encoder Net $g_o$ & MLP(obstacle\_size,32), MLP(32,32) \\
    \hline
    \multirow{3}[2]{*}{Attention Net} & Key Network $f_{K_{(\cdot)}^{(\cdot)}}$: MLP(32,32) \\
\cline{2-2}           & Query Network $f_{Q_{(\cdot)}^{(\cdot)}}$: MLP(32,32) \\
\cline{2-2}           & Value Network $f_{V_{(\cdot)}^{(\cdot)}}$: MLP(32,32) \\
    \hline
    Feedforward Net & MLP(32,32),MLP(32,32) \\
    \hline
    Node Message Passing $f_x$ & MLP(32*4,32),MLP(32,32) \\
    \hline
    Edge Message Passing $f_y$ & MLP(32*3,32),MLP(32,32) \\
    \hline    
    \multicolumn{2}{c}{\textbf{Stage2 Local Planner}} \\
    \hline
    \multirow{2}[2]{*}{Planner Net $f_P$} & MLP(32+obstacle\_size*window\_size, 64),MLP(64,32), \\
\cline{2-2}           & MLP(32,32), MLP(32,1) \\
    \hline
    \end{tabular}%
  \label{tab:netdetail}%
\end{table}%

\section{Experiments}

\subsection{Hyperparameters}
We provide the hyperparameters in Table \ref{tab:hyper}.
\begin{table}[H]
  \centering
  \caption{Hyperparameters}
    \begin{tabular}{c|c}
    \toprule[2pt]
    \textbf{Hyperparameters} & \textbf{Values} \\
    \hline
    $k$ for k-NN & 50 \\
    \hline
    Training Epoch before DAgger & 200 \\
    \hline
    Training Epoch for DAgger & 100 \\
    \hline
    Learning Rate & 1e-3 \\
    \hline
    Temporal Encoding Frequency $\omega$ & 10000 \\
    \hline
    $d_{TE}$ & 32 \\
    \hline
    Time Window $w$ & 2 \\
    \hline
    \end{tabular}%
  \label{tab:hyper}%
\end{table}%

\subsection{Overall Performance}
We provide the detailed overall performance in Table \ref{tab:suc}, \ref{tab:pr} and \ref{tab:cc}.

\begin{table}[H]  
  \centering
\caption{Success Rate (\%)}
\begin{adjustbox}{width=1.2\columnwidth,center}
    \begin{tabular}{c|c|c|c|c|c|c|c}
    \toprule[2pt]
           &        & \textbf{2Arms} & \textbf{Kuka-4DoF} & \textbf{Kuka-5DoF} & \textbf{Kuka-7DoF} & \textbf{3Arms} & \textbf{Kuka3Arms} \\
    \hline
    \multirow{2}[1]{*}{\textbf{SIPP}} & random & 100±0.00 & 100±0.00 & 100±0.00 & 100±0.00 & 100±0.00 & 100±0.00 \\
\cline{2-8}           & hard   & 100±0.00 & 100±0.00 & 100±0.00 & 100±0.00 & 100±0.00 & 100±0.00 \\
    \hline
    \multirow{2}[1]{*}{\textbf{GNN-TE}} & random & \textbf{94.1±0.02} & \textbf{97.8±0.01} & 97.6±0.00 & \textbf{98.8±0.01} & \textbf{92.1±0.01} & \textbf{97.4±0.01} \\
\cline{2-8}           & hard   & \textbf{62.5±0.02} & \textbf{34.9±0.00} & \textbf{37.9±0.15} & \textbf{38.1±0.12} & \textbf{52.5±0.03} & 42.1±0.08 \\
    \hline
    \multirow{2}[1]{*}{\textbf{GNN-TE w/o Dagger}} & random & \textbf{94.1±0.01} & 97.5±0.01 & \textbf{97.7±0.00} & \textbf{98.8±0.01} & 91.5±0.01 & 97.3±0.01 \\
\cline{2-8}           & hard   & 58.1±0.03 & 33.3±0.00 & 36.8±0.14 & 36.5±0.11 & 51.6±0.05 & \textbf{42.6±0.08} \\
    \hline
    \multirow{2}[1]{*}{\textbf{Dijkstra-H}} & random & 89.7±0.03 & 96.3±0.01 & 96.2±0.01 & 97.7±0.01 & 85.9±0.01 & 93.9±0.01 \\
\cline{2-8}           & hard   & 0.00±0.00      & 0.00±0.00     & 0.00±0.00     & 0.00±0.00     & 0.00±0.00   & 0.00±0.00 \\
    \hline
    \end{tabular}%
  \label{tab:suc}%
  \end{adjustbox}
\end{table}%

\begin{table}[H]
  \centering
  \caption{Path Time Ratio}
  \begin{adjustbox}{width=1.2\columnwidth,center}
    \begin{tabular}{c|c|c|c|c|c|c|c}
    \toprule[2pt]
           &        & \textbf{2Arms} & \textbf{Kuka-4DoF} & \textbf{Kuka-5DoF} & \textbf{Kuka-7DoF} & \textbf{3Arms} & \textbf{Kuka3Arms} \\
    \hline
    \multirow{2}[1]{*}{\textbf{SIPP}} & random & 100±0.00 & 100±0.00 & 100±0.00 & 100±0.00 & 100±0.00 & 100±0.00 \\
\cline{2-8}           & hard   & 100±0.00 & 100±0.00 & 100±0.00 & 100±0.00 & 100±0.00 & 100±0.00 \\
    \hline
    \multirow{2}[1]{*}{\textbf{GNN-TE}} & random & \textbf{107.55±2.33} & \textbf{132.71±4.46} & 171.39±12.94 & 172.83±8.77 & \textbf{120.76±7.89} & 186.18±23.96 \\
\cline{2-8}           & hard   & 123.31±7.71 & \textbf{189.54±5.41} & \textbf{183.33±48.26} & \textbf{167.65±47.7} & \textbf{134.54±9.76} & \textbf{152.52±22.31} \\
    \hline
    \multirow{2}[1]{*}{\textbf{GNN-TE w/o Dagger}} & random & 109.67±5.46 & 148.1±13.59 & \textbf{161.40±16.50} & \textbf{170.05±9.11} & 130.48±11.74 & \textbf{181.74±18.98} \\
\cline{2-8}           & hard   & \textbf{121.41±8.85} & 250.26±82.5 & 232.67±99.79 & 193.12±42.4 & 154.85±23.81 & 164.67±40.65 \\
    \hline
    \multirow{2}[1]{*}{\textbf{Dijkstra-H}} & random & 123.73±8.29 & 212.09±20.73 & 198.72±14.7 & 199.22±23.22 & 177.88±19.59 & 189.23±5.64\\
\cline{2-8}           & hard   & /      & /      & /      & /      & /      & / \\
    \hline
    \end{tabular}%
  \label{tab:pr}%
    \end{adjustbox}
\end{table}%

\begin{table}[H]
  \centering
  \caption{Collision Checking}
  \begin{adjustbox}{width=1.2\columnwidth,center}
    \begin{tabular}{c|c|c|c|c|c|c|c}
    \toprule[2pt]
           &        & \textbf{2Arms} & \textbf{Kuka-4DoF} & \textbf{Kuka-5DoF} & \textbf{Kuka-7DoF} & \textbf{3Arms} & \textbf{Kuka3Arms} \\
    \hline
    \multirow{2}[1]{*}{\textbf{SIPP}} & random & 60440.21±1543.21 & 171336.68±2061.60 & 196567.99±1152.81 & 268602.98±780.07 & 96713.81±3945.07 & 269033.61±1159.78 \\
\cline{2-8}           & hard   & 1080768.34±81176.44 & 145280.09±1448.0 & 182696.61±1271.86 & 257783.45±742.83 & 114337.00±3560.95 & 255173.7±2099.46 \\
    \hline
    \multirow{2}[1]{*}{\textbf{GNN-TE}} & random & \textbf{17.24±8.45} & \textbf{37.36±4.74} & \textbf{56.89±5.41} & \textbf{61.99±8.08} & \textbf{33.08±7.95} & \textbf{72.63±13.13} \\
\cline{2-8}           & hard   & \textbf{47.31±7.72} & \textbf{155.7±96.49} & \textbf{108.25±48.29} & \textbf{110.65±39.63} & \textbf{65.93±14.84} & \textbf{90.42±25.61} \\
    \hline
    \multirow{2}[1]{*}{\textbf{GNN-TE w/o Dagger}} & random & 21.13±11.71 & 42.00±4.60 & 61.79±10.39 & 69.39±10.67 & 47.48±14.40 & 73.33±11.57 \\
\cline{2-8}           & hard   & 47.91±8.85 & 160.43±70.64 & 166.61±86.29 & 164.41±88.98 & 98.67±37.69 & 98.41±32.15 \\
    \hline
    \multirow{2}[1]{*}{\textbf{Dijkstra-H}} & random & 56.21±23.56 & 236.50±17.60 & 229.35±21.66 & 236.95±34.93 & 103.74±19.57 & 237.93±12.83 \\
\cline{2-8}           & hard   & /      & /      & /      & /      & /      & / \\
    \hline
    \end{tabular}%
  \label{tab:cc}%
  \end{adjustbox}
\end{table}%

\subsection{Optional Backtracking Search}\label{AppendixBT}
We provide the results of \textbf{GNN-TE} and \textbf{Dijkstra-H} with backtracking (top-5) in 2Arms environment in \ref{backtracking-with-dij}. Our method outperforms the heuristic method with and without the backtracking search.
\begin{table}[h!]
  \centering
  \caption{The performance of backtracking search in the 2Arms environment}
  \begin{adjustbox}{center}
    \begin{tabular}{c|c|c||c|c||c|c}
    \toprule[2pt]
           &        & \textbf{SIPP} & \textbf{Dijkstra-H} &  \textbf{GNN-TE}& \textbf{Dijkstra-H w. BT} & \textbf{GNN-TE w. BT} \\
    \hline
    \multirow{1}[4]{*}{{\textbf{Success Rate}}} & random & 100\%  & 89.70\% & \textbf{94.10}\% & 94.10\% & \textbf{98.00\%}\\
\cline{2-7}           & hard   & 100\%  & 0\% & \textbf{62.50}\% & 50.70\% & \textbf{89.30\%} \\
    \hline
    \multirow{1}[4]{*}{\textbf{Path Time Ratio}} & random & 100\%  & 123.61\% & \textbf{107.65}\% & 123.61\% & \textbf{107.65\%} \\
\cline{2-7}           & hard   & 100\%  & /      & \textbf{128.22}\% & 276.25\% & \textbf{128.22\%} \\
    \hline
    \multirow{1}[4]{*}{\textbf{Collision Checks}} & random & 60K    & 55.88  & \textbf{17.17}  & 55.88 & \textbf{17.17} \\
\cline{2-7}           & hard   & 1081K  & /      & \textbf{52.68}  & 1161.29 & \textbf{52.68} \\
    \hline
    \end{tabular}%
  \label{backtracking-with-dij}%
  \end{adjustbox}
\end{table}%

\textcolor{black}{We also provide the success rate of \textbf{GNN-TE} with backtracking in all the environments in \ref{tab:backtracking-all}. As the DoF and the complexity of the configuration space increase, the searching space grows and requires more backtracking steps. Thus the increase in success rate by backtracking may not be as significant as in the simple settings if we keep the backtracking steps the same. However, GNN-TE still shows a significant advantage over Dijkstra-H even with backtracking in all the settings.}

\begin{table}[H]
  \centering
  \caption{Success rates of GNN-TE and Dijkstra-H with backtracking search}
  \begin{adjustbox}{center}
    \begin{tabular}{c|c|c|c|c|c|c|c}
    \toprule[2pt]
           &        & \textbf{2Arms} & \textbf{Kuka-4DoF} & \textbf{Kuka-5DoF} & \textbf{Kuka-7DoF} & \textbf{3Arms} & \textbf{Kuka3Arms} \\
    \hline 
    \multirow{4}[1]{*}{\textbf{random}} & \textbf{Dijkstra-H} & 89.7±0.03 & 96.3±0.01 & 96.2±0.01 & 97.7±0.01 & 85.9±0.01 & 93.9±0.01 \\
\cline{2-8}           & \textbf{GNN-TE} & \textbf{94.1±0.02} & \textbf{97.8±0.01} & \textbf{97.6±0.00} & \textbf{98.8±0.01} & \textbf{92.1±0.01} & \textbf{97.4±0.01} \\
\cline{2-8}    & \textbf{Dijkstra-H w. BT} & 94.1±0.01 & 96.7±0.02 & 96.2±0.01 & 97.8±0.01 & 92.4±0.01 & 94.2±0.01 \\
\cline{2-8}           & \textbf{GNN-TE w. BT} & \textbf{98.0±0.01} & \textbf{97.8±0.01} & \textbf{97.7±0.00} & \textbf{98.9±0.01} & \textbf{97.1±0.01} & \textbf{97.4±0.00} \\
    \hline \hline
    \multirow{4}[1]{*}{\textbf{hard}} & \textbf{Dijkstra-H} & 0.0±0.0 & 0.0±0.0 & 0.0±0.0 & 0.0±0.0 & 0.0±0.0 & 0.0±0.0 \\
\cline{2-8}           & \textbf{GNN-TE} & \textbf{62.5±0.02} & \textbf{34.9±0.00} & \textbf{37.9±0.15} & \textbf{38.1±0.12} & \textbf{52.5±0.03} & \textbf{42.1±0.08} \\
\cline{2-8}           & \textbf{Dijkstra-H w. BT} & 50.7±0.06 & 10.1±0.01 & 5.8±0.40 & 2.8±0.24 & 45.8±0.05 & 2.7±0.01 \\
\cline{2-8}           & \textbf{GNN-TE w. BT} & \textbf{89.3±0.03} & \textbf{36.4±0.00} & \textbf{40.8±0.15} & \textbf{39.4±0.12} & \textbf{82.6±0.02} & \textbf{44.3±0.01} \\
    \hline
    \end{tabular}%
  \label{tab:backtracking-all}%
  \end{adjustbox}
\end{table}%

\subsection{Comparison with End-to-End RL}\label{AppendixRL}
We compare our approach with RL-based approaches, \textbf{DQN-GNN} and \textbf{PPO-GNN} specifically. The two algorithms both encode the graph using GNN as ours in stage 1. \textbf{DQN-GNN}, similar to our local planner, learns a network to evaluate the Q value of the subsequent edge as a priority value. \textbf{PPO-GNN} learns a policy network that output the next configuration, and we project it onto the nearest vertex on the graph encoded by GNN. We define the reward as $-10$ for collision, $10$ for reaching the goal, and the $distance\ displacement$ for non-collision configurations.

In the general RL setting, we do not expect the generalization capability of algorithms. But as a general graph encoder, GNN can achieve generalization between graphs. Based on this, we train \textbf{DQN-GNN} and \textbf{PPO-GNN} across problems and test their performance. On training set, \textbf{DQN-GNN} achieves $54.5\%$ success rate while \textbf{PPO-GNN} only achieves $21.1\%$. We also provide results on randomly generated test cases and hard cases in Table \ref{table:RL}. We can observe that \textbf{GNN-TE} significantly outperforms all the RL approaches. Moreover, the advanced inductive bias of GNN for discrete decision-making problems explained why \textbf{DQN-GNN} has better performance than \textbf{PPO-GNN} in both randomly sampled cases and hard cases. Nevertheless, \textbf{DQN-GNN} and \textbf{PPO-GNN} both cannot efficiently find plans, especially in hard cases. This is because RL-based approaches have trouble finding a feasible path without demonstration from the oracle and only rely on rewards in challenging problems. 

\vspace{-10pt}

\begin{table}[!ht]
  \centering
  \caption{Table for RL Approaches in 2Arms Environment}
    \begin{tabular}{c|c|c|c|c}
    \toprule[2pt]
           &        & \textbf{GNN-TE} & \textbf{DQN-GNN} & \textbf{PPO-GNN} \\
    \hline
    \multirow{2}[1]{*}{\textbf{Success Rate}} & random & \textbf{94.10\%} & 62.40\% & 9.80\% \\
\cline{2-5}           & hard   & \textbf{62.50\%} & 2.00\% & 0.70\% \\
    \hline
    \multirow{2}[1]{*}{\textbf{Path Time Ratio}} & random & \textbf{103.55\%} & 105.47\% & 119.73\% \\
\cline{2-5}           & hard   & \textbf{102.43\%} & 109.76\% & 134.15\% \\
    \hline
    \multirow{2}[1]{*}{\textbf{Collision Checking}} & random & \textbf{4.98} & \textbf{4.68} & 5.43 \\
\cline{2-5}           & hard   & \textbf{6.00} & \textbf{6.00} & 7.00 \\
    \hline
    \end{tabular}%
  \label{table:RL}%
\end{table}%

\subsection{Comparison with OracleNet-D}\label{AppendixOracle}
We compare \textbf{GNN-TE} with a learning-based approach \textbf{OracleNet-D} by modifying \textbf{OracleNet} \cite{OracleNet} to the dynamic version. Concretely, we concatenate the trajectories of obstacles to the input in every roll-out of \textbf{OracleNet} to inform the network of the dynamic environment \footnote{We use the original code from repository https://github.com/mayurj747/oraclenet-analysis}. We provide the results in 2Arms environment in Table \ref{tab:Oracle}. (For a fair comparison, we present the result of GNN-TE without DAgger. And the collision checking is not provided because OracleNet-D generates and rolls out the path iteratively without checking the collision.)

\begin{table}[H]
  \centering
  \caption{Table for GNN-TE and OracleNet-D in 2Arms Environment}
    \begin{tabular}{c|c|c|c|c|c}
    \toprule[2pt]
           &        & \textbf{SIPP} & \textbf{Dijkstra-H} & \textbf{GNN-TE} & \textbf{OracleNet-D} \\
    \hline
    \multirow{2}[1]{*}{\textbf{Success Rate}} & random & 100\%  & 89.70\% & \textbf{94.10\%} & 53.90\% \\
\cline{2-6}           & hard   & 100\%  & 0.00\% & \textbf{58.10\%} & 10.80\% \\
    \hline
    \multirow{2}[1]{*}{\textbf{Avg Path Time Ratio}} & random & 100\%  & 120.61\% & \textbf{113.94\%} & 1130.76\% \\
\cline{2-6}           & hard   & 100\%  & /      & \textbf{118.92\%} & 813.00\% \\
    \hline
    \end{tabular}%
  \label{tab:Oracle}%
\end{table}%

We observe that the performance of \textbf{OracleNet-D} falls behind \textbf{GNN-TE} both on success rate and the average time ratio. This result shows that encoding environmental information is important for the planner in a dynamic environment. As mentioned in \cite{OracleNet}, the configuration of the robot and the environmental information form different distributions and the mapping is challenging. We believe GNN with the attention mechanism and temporal encoding provides a good solution to the problem. Also, GNN-TE benefits from the second-stage local planner, which takes local temporal obstacle information into consideration.

\subsection{\textcolor{black}{Ablation Study on Varying Training Set Sizes}}
\begin{figure}[H]
  \centering
  \begin{adjustbox}{width=0.7\textwidth,center=\textwidth}
    \includegraphics[]{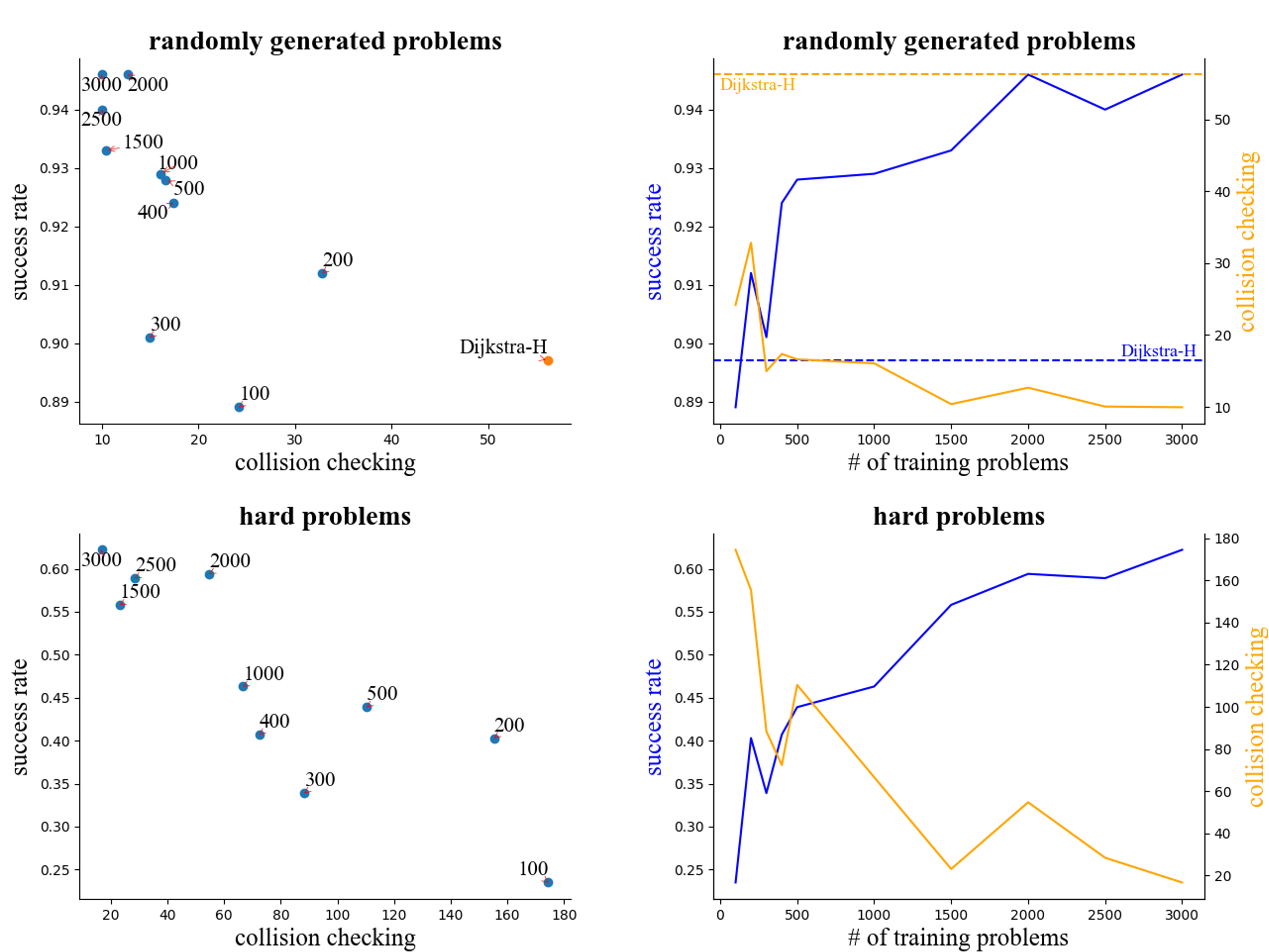}
    \end{adjustbox}
    \caption{Results on varying training set sizes in 2Arms environment. We observe that \textbf{GNN-TE} benefits from increasing the training problems, both in better success rate and less collision checking. \textbf{Left:} A scatter plot visualizes the relevance between the success rate and the collision checking regarding the training size. The number on each point indicates the training size. \textbf{Right:} An equivalent plot that clearly shows the performance boost benefited from a larger training size. Higher success rate (blue curve) and lower collision checking (orange curve) are favored.}
    \label{fig:tradeoff}
\end{figure} 
 We train \textbf{GNN-TE} on varying training problems (specifically 100, 200, 300, 400, 500, 1000, 1500, 2000, 2500, 3000) and test on the same random sampled and hard problems in 2Arms environment.
 
 We observe that \textbf{GNN-TE} benefits from increasing the training problems, both in better success rate and less collision checking. From the plot in the right column of the figure, we observed that the trends are prone to be log-like. It shows that the performance will be saturated as the training set covers the problem distribution.

\subsection{\textcolor{black}{Ablation Study on Basic GNN}}
In Table \ref{tab:basicGNN}, we provide the overall performance gain by all the components of \textbf{GNN-TE} over the basic GNN (\textbf{GNN-basic}) in 2Arms environment. Specifically, in the first stage, \textbf{GNN-basic} removes the attention mechanism and temporal encoding. And in the second stage, \textbf{GNN-basic} only inputs the obstacle encoding at the current time step.

\begin{table}[H]
  \centering
  \caption{Overall performance gain over basic GNN}
    \begin{tabular}{c|c|c|c|c|c}
    \toprule[2pt]
           &        & \textbf{SIPP} & \textbf{Dijkstra-H} & \textbf{GNN-TE} & \textbf{GNN-basic} \\
    \hline
    \multirow{2}[1]{*}{\textbf{Success Rate}} & random & 100\%  & 89.70\% & \textbf{94.10\%} & 92.70\% \\
\cline{2-6}           & hard   & 100\%  & 0.00\% & \textbf{62.50\%} & 32.00\% \\
    \hline
    \multirow{2}[1]{*}{\textbf{Avg Path Time Ratio}} & random & 100\%  & 123.73\% & \textbf{107.78\%} & 112.42\% \\
\cline{2-6}           & hard   & 100\%  & /      & \textbf{122.13\%} & 185.92\% \\
    \hline
    \multirow{2}[1]{*}{\textbf{Avg Collision Checking}} & random & 60K    & 56.21  & \textbf{17.44} & 28.80 \\
\cline{2-6}           & hard   & 1081K  & /      & \textbf{45.23} & 109.70 \\
    \hline
    \end{tabular}%
  \label{tab:basicGNN}%
\end{table}%

\subsection{\textcolor{black}{Failure Modes in 2Arms Environment}}\label{failmode}
\begin{figure}[H]
  \centering
  \begin{adjustbox}{width=0.6\textwidth,center=\textwidth}
    \includegraphics[]{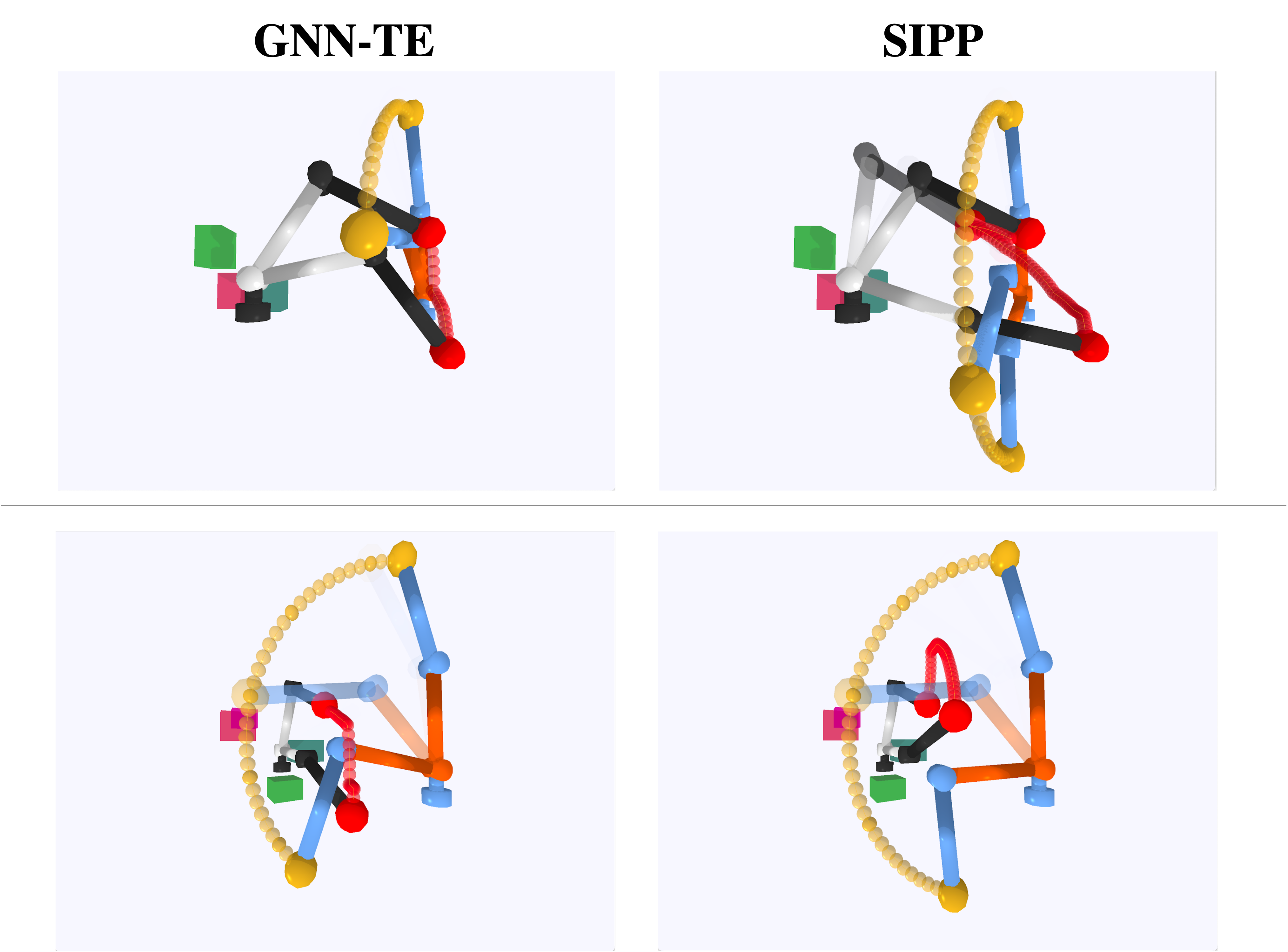}
    \end{adjustbox}
    \caption{Failure mode: the planner fails to make a detour. Our planner controls the arm in black and white.}
    \label{fig:fm1}
\end{figure}

\begin{figure}[H]
  \centering
  \begin{adjustbox}{width=0.6\textwidth,center=\textwidth}
    \includegraphics[]{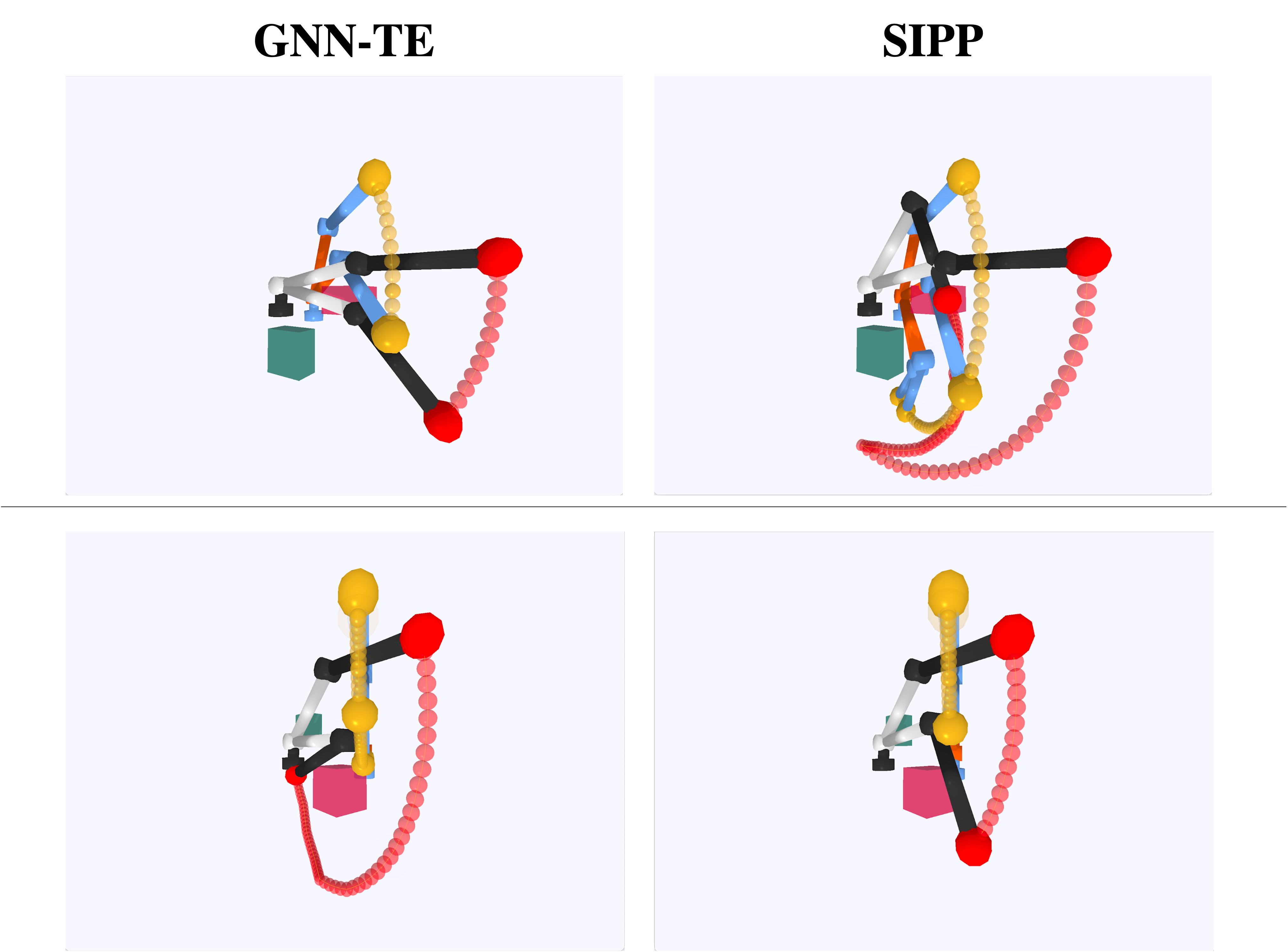}
    \end{adjustbox}
    \caption{Failure mode: the planner gets too close to the obstacle. Our planner controls the arm in black and white. Though the planner follows the correct direction, it gets too close to the obstacle arm, which leads to the collision.}
    \label{fig:fm2}
\end{figure}

We provide visualizations of \textbf{GNN-TE} failing to find feasible solutions in 2Arms environments. We find there are mainly two modes: it fails to make a detour in Fig. \ref{fig:fm1} or gets too close to the moving obstacles in Fig.\ref{fig:fm2}. In Fig. \ref{fig:fm1}, we can observe that \textbf{GNN-TE} plans to directly get to the goal while the feasible path is to make a detour to avoid the obstacle. In Fig. \ref{fig:fm2}, although \textbf{GNN-TE} can follow the correct direction but fail in getting too close to the obstacle arm.

\section{\textcolor{black}{Limitations and Future Work}}
\subsection{Discussions on Using GNNs and Attention Mechanism}
Motion planning has been a longstanding challenge in robotics, especially in dynamic environments. Our approach uses a learning-based approach leveraging Graph Neural Networks to efficiently tackle this problem. GNNs show great capability in capturing geometric information and are invariant to the permutations of the sampled graph. Another challenge in dynamic environment is that the difference in distributions of the robot configuration and the environmental information makes the mapping and motion planning challenging. Our approach tackles this by introducing the attention mechanism with temporal encoding to learn the correlation between the temporal positions of obstacles and the ego-arm configuration on the graph. It is efficient because, as for a configuration node on the graph, the obstacles' positions of some time steps are more important than others, as the obstacles may have more possibilities of colliding with the ego-arm at those time steps. So, in this case, the obstacles of those time steps should be given more importance in modeling. Also, the attention mechanism can take time sequences with variable length as inputs.

Regardless of the empirical efficiency, the performance of the GNN-based approach is still bounded by the sampled configurations. It can only be boosted by a sufficient number of nodes on the graph, especially in a complex environment and with a robot with a high degree of freedom. It is still an open problem how many samples would be sufficient for the GNN to capture the geometric pattern from the configuration space.

\subsection{Limited Performances on Hard Problems}% or Limitations for Inference on Hard Problems
Although \textbf{GNN-TE} can achieve a better success rate than other learning-based approaches, it is still not complete and has limited performance on hard problems (see Section \ref{failmode} for examples of failure modes). A direction of solving this problem is to do hard example mining and train on those problems, where we train \textbf{GNN-TE} on extra hard examples and test its performance, and the success rate rises from 62.5\% to 71.3\% on 2Arms environment. However, in general, we believe the safety and reliability of learning-enabled systems are always a core issue that needs to be solved after learning-based approaches show clear benefits. 

For motion planning, a potential future direction is to integrate our learning-based component with monitoring. Such monitoring identifies hard graph structures that are out-of-distribution for the neural network components. It ensures that the learning-based components are only used when the planning can be safely accelerated, in which case they will provide great benefits in reducing collision checking and overall computation. When hard or out-of-distribution cases occur, the planner should fall back to more complete algorithms such as \textbf{SIPP}. There also has been much ongoing development in frameworks for ensuring the safe use of learning-based components in planning and control, which we believe is orthogonal to our current work. For example, \cite{brunke2022safe} provides reviews learning-based control and RL approaches that prioritize the safety of the robot’s behavior.

\subsection{Trade-off Between Quality and Efficiency}
Another observation from the result is the trade-off between quality (success rate of finding paths) and efficiency (number of collision checks). In this work, we further add backtracking, where we keep a stack of policy edges of the top-n priority values and allow the algorithm to take the sub-optimal choices if it fails. Therefore, the backtracking will increase the collision checking with the hope of finding a solution. Although adding this or other systematic searching algorithms can improve the quality in the sacrifice of efficiency, we think the actual bottleneck might still be the priority values as the heuristic produced by the model. We believe this trade-off may be a crucial learning-based dynamic motion planning topic and needs future investigations.

\subsection{Problem Distribution and Generalization}
As most learning-based approaches would assume, our model needs to be trained on the same actor and obstacle arms as it's tested on. Both the sampled graph and the expert trajectory are implicitly conditioned on the kinematic structure. This assumption aligns with the most immediate use of learning-based components for reducing repeated planning computation in a relatively fixed setting of arm configurations. We believe learning planning models that can be generalized to arbitrary arms and obstacles is still challenging for the community, for it requires an in-depth study of other issues that have not been fully understood, such as the inherent generalization properties of graph neural networks. As shown in \cite{garg2020generalization}, there still exists the trade-off between expressivity and generalization in GNN. We leave this topic to future works.

\section{More Snapshots in Different Environments}
In this section, we show more snapshots of baselines SIPP, GNN-TE, Dijkstra-H in different environments. In those environments and cases, \textbf{GNN-TE} succeeds in finding a near-optimal path while \textbf{Dijkstra-H} fails.

\end{appendices}

\begin{figure}[H]
  \centering
  \begin{adjustbox}{width=0.9\textwidth,center=\textwidth}
    \includegraphics[]{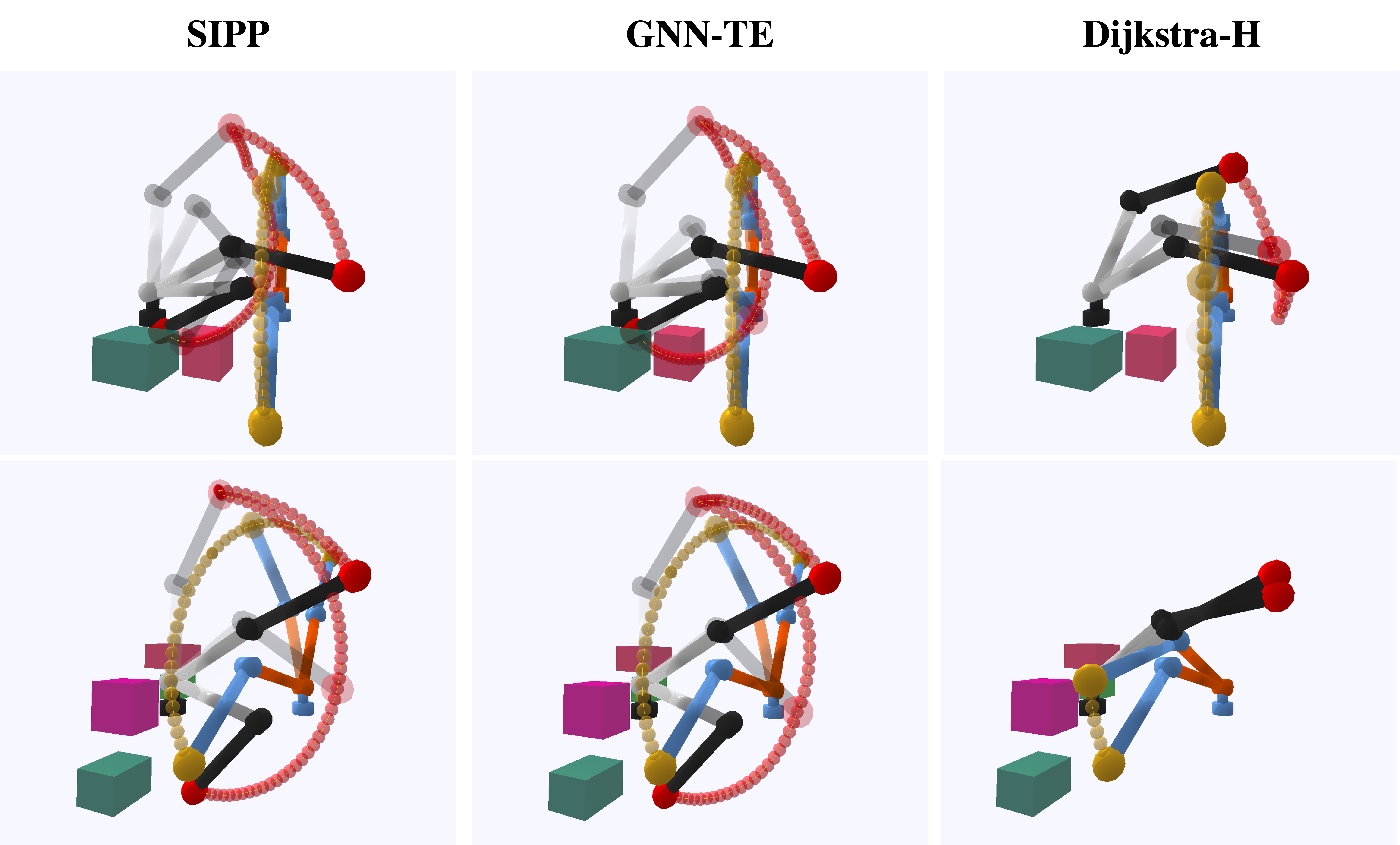}
    \end{adjustbox}
    \caption{Snapshots: 2Arms}
\end{figure} 

\begin{figure}[H]
  \centering
  \begin{adjustbox}{width=0.9\textwidth,center=\textwidth}
    \includegraphics[]{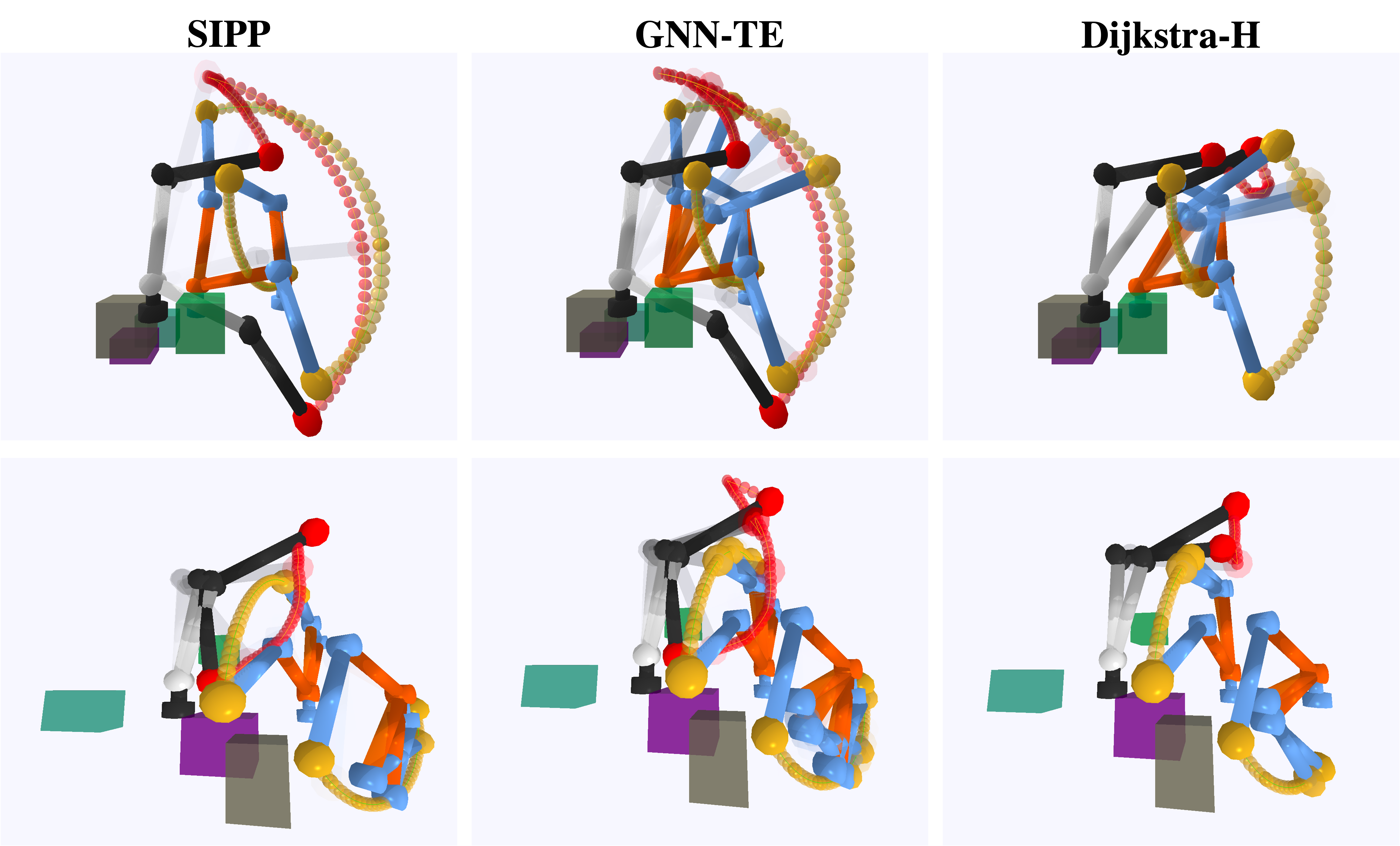}
    \end{adjustbox}
    \caption{Snapshots: 3Arms}
\end{figure} 

\begin{figure}[H]
  \centering
  \begin{adjustbox}{width=0.9\textwidth,center=\textwidth}
    \includegraphics[]{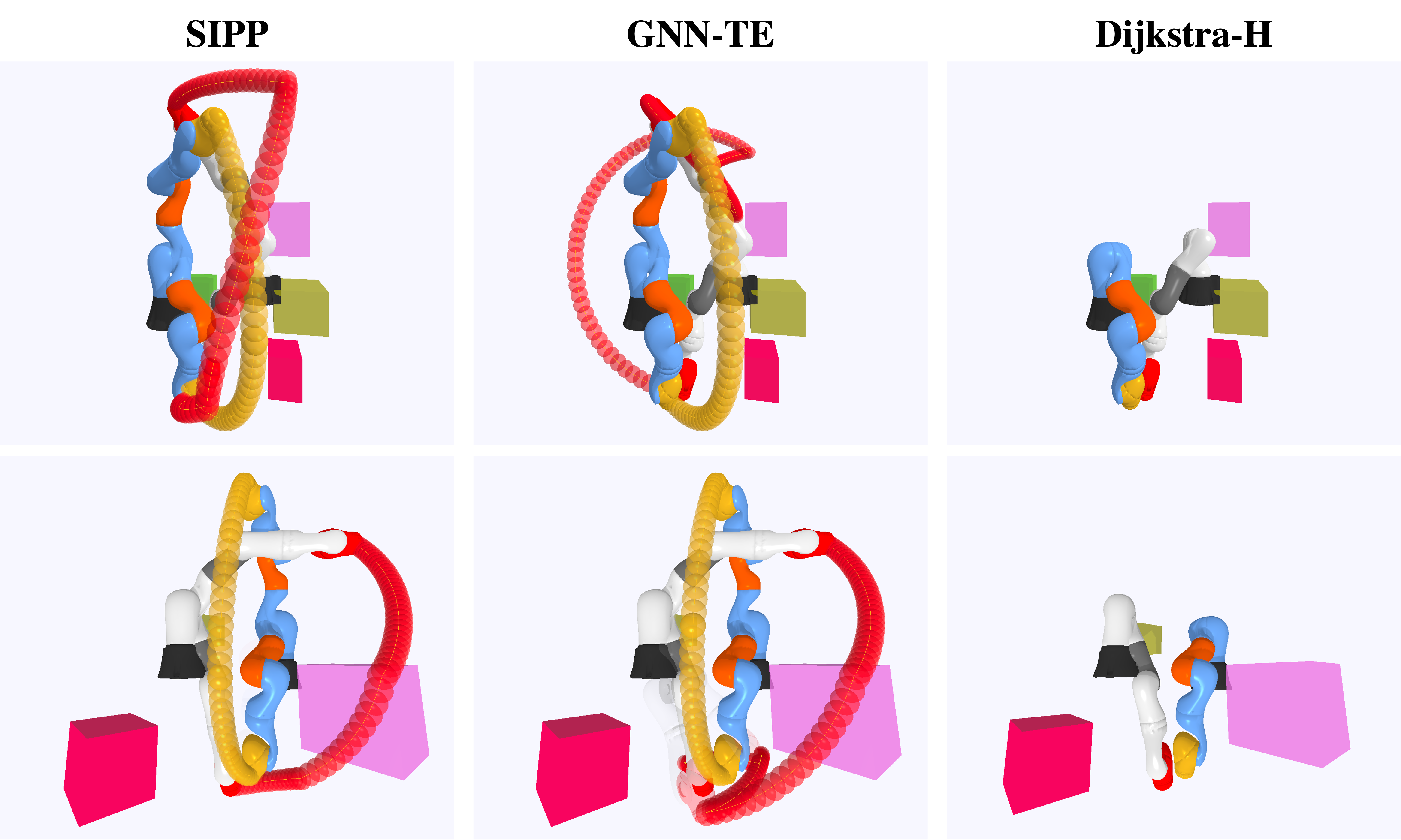}
    \end{adjustbox}
    \caption{Snapshots: Kuka-4DoF}
\end{figure} 

\begin{figure}[H]
  \centering
  \begin{adjustbox}{width=0.9\textwidth,center=\textwidth}
    \includegraphics[]{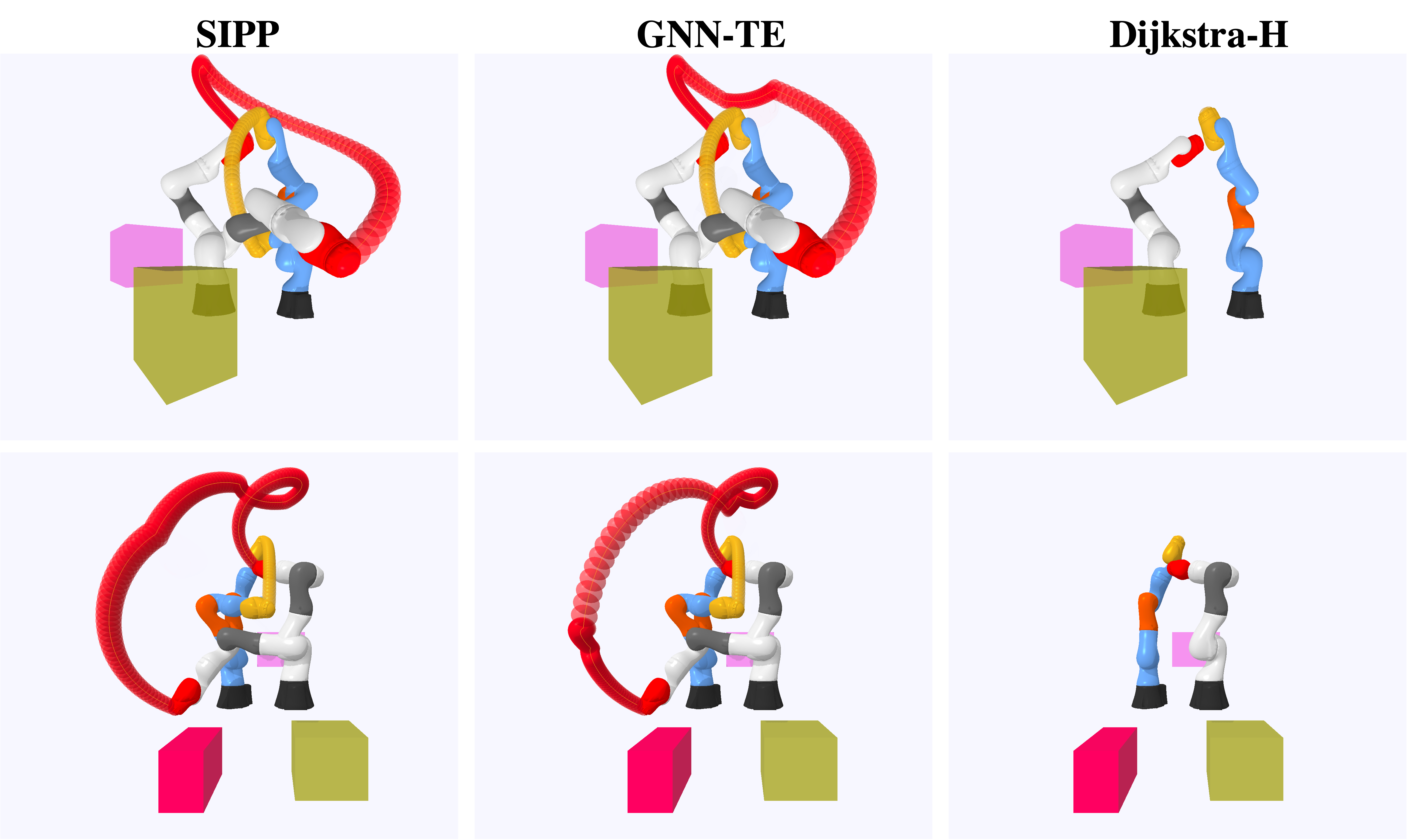}
    \end{adjustbox}
    \caption{Snapshots: Kuka-5DoF}
\end{figure} 

\begin{figure}[H]
  \centering
  \begin{adjustbox}{width=0.9\textwidth,center=\textwidth}
    \includegraphics[]{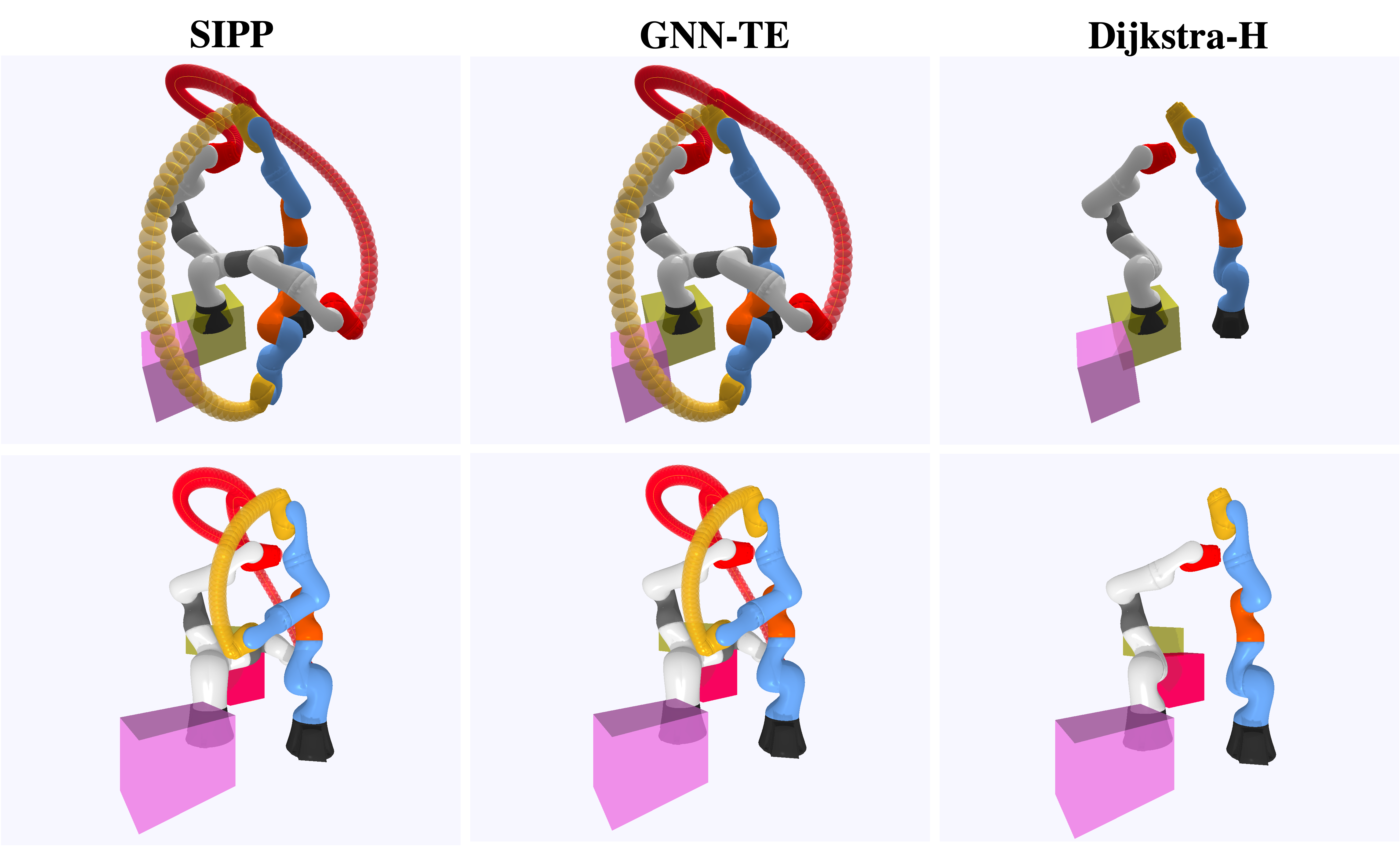}
    \end{adjustbox}
    \caption{Snapshots: Kuka-7DoF}
\end{figure} 

\begin{figure}[H]
  \centering
  \begin{adjustbox}{width=0.9\textwidth,center=\textwidth}
    \includegraphics[]{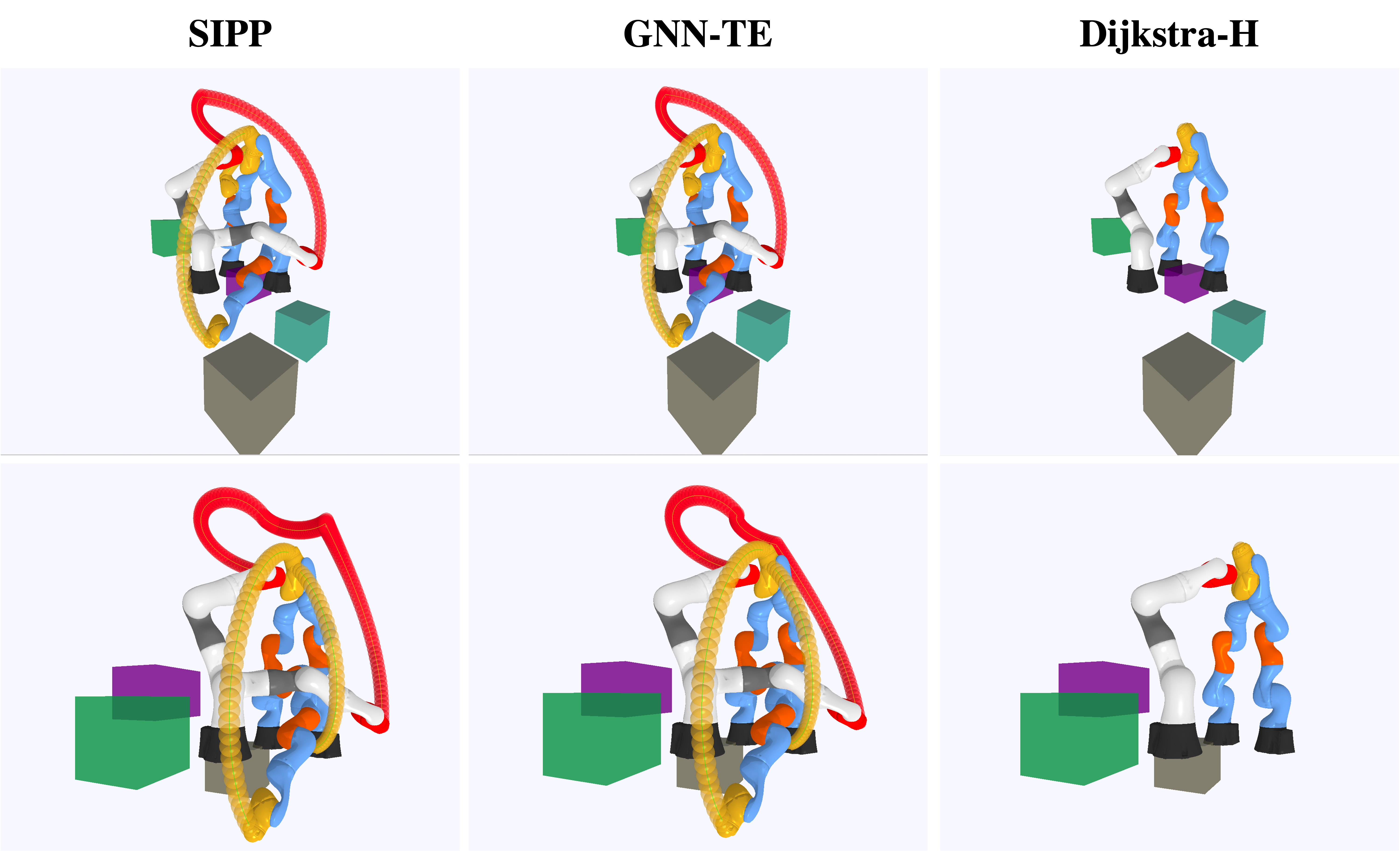}
    \end{adjustbox}
    \caption{Snapshots: Kuka3Arms}
\end{figure}

\end{document}